\documentclass[letterpaper, 10 pt, journal, twoside]{IEEEtran}
\IEEEoverridecommandlockouts
\usepackage{soul}

\usepackage{cite}
\usepackage{textcomp}
\usepackage{amsmath,amssymb,amsfonts}
\usepackage{contour}

\usepackage{fontawesome}
\usepackage{colortbl}
\usepackage{wrapfig}
\usepackage{graphicx}

\usepackage{hyperref}
\hypersetup{colorlinks,linkcolor=black, citecolor=black}
\usepackage{subfig}
\usepackage{caption}
\usepackage{float}
\usepackage{romannum}
\usepackage[T1]{fontenc}
\usepackage{fontawesome}

\usepackage{multirow}
\usepackage{booktabs}
\usepackage{tcolorbox}
\usepackage{makecell}
\usepackage{hhline}
\usepackage{scalerel}

\usepackage{footnote}
\makesavenoteenv{tabular}

\usepackage{array}
\usepackage{xspace}
\usepackage{bigstrut}
\usepackage[export]{adjustbox}
\usepackage{ragged2e}
\usepackage[linesnumbered, ruled,vlined]{algorithm2e}

\usepackage{stfloats}
\usepackage{paralist}
\usepackage{framed}

\usepackage{listings}

\usepackage{stfloats}

\usepackage{boldline}

\definecolor{gray05}{gray}{0.95}

\newcommand{\ind}{\perp\!\!\!\!\perp} 

\SetCommentSty{mycommfont}
\newcommand{\texTTT}{\fontfamily{lmtt}\selectfont}

\newcommand{\orcid}[1]{\href{https://orcid.org/#1}{\includegraphics[width=10pt]{images/orcid.png}}}
\DeclareMathAlphabet{\mathpzc}{OT1}{pzc}{m}{it}
\usepackage{tikz}
\newcommand*\circled[1]{\tikz[baseline=(char.base)]{
            \node[shape=circle,draw,inner sep=1pt] (char) {#1};}}
\usetikzlibrary{arrows.meta}

\newcommand\edgeone{
\begin{tikzpicture}
  \draw[black,  arrows={-Triangle[angle=90:3pt,black,fill=black]}] (0,0.0) -- (0.5,0.0);
\end{tikzpicture}\xspace}

\newcommand\edgetwo{
\begin{tikzpicture}
  \draw[black,  arrows={Triangle[angle=90:3pt,black,fill=black]-Triangle[angle=90:3pt,black,fill=black]}] (0,0.0) -- (0.5,0.0);   
\end{tikzpicture}\xspace}

\newcommand\edgethree{
\begin{tikzpicture}
  \draw[black,  arrows={Circle[open]-Triangle[angle=90:3pt,black,fill=black]}] (0,0.0) -- (0.5,0.0);   
\end{tikzpicture}\xspace}

\newcommand\edgefour{
\begin{tikzpicture}
  \draw[black,  arrows={Circle[open]-Circle[open]}] (0,0.0) -- (0.5,0.0);   
\end{tikzpicture}\xspace}

\usepackage{MnSymbol}
\usepackage{amsmath}
\DeclareMathOperator*{\argmax}{arg\,max}

\def\hat{\widehat}

\begin{document}
\bstctlcite{IEEEexample:BSTcontrol}

\title{\textsc{CaRE}: Finding Root Causes of Configuration Issues in Highly-Configurable Robots\\
}

\author{Md Abir Hossen$^{1}$, Sonam Kharade$^{1}$, Bradley Schmerl$^{2}$, Javier Cámara$^{3}$, Jason M. O'Kane$^{4}$,\\ Ellen C. Czaplinski$^{5}$, Katherine A. Dzurilla$^{5}$, David Garlan$^{2}$ and Pooyan Jamshidi$^{1}$% <-this % stops a space

%Journal requirement
\thanks{This work was supported by National Aeronautics and Space Administration~(Award 80NSSC20K1720) and  National Science Foundation~(Award 2107463).}

\thanks{$^{1}$M. A. Hossen, S. Kharade, and P. Jamshidi are with College of Engineering and Computing, University of South Carolina, SC, USA (e-mail: abir.hossen786@gmail.com; skharade@mailbox.sc.edu; pjamshid@cse.sc.edu)}
        
\thanks{$^{2}$B. Schmerl, and D. Garlan are with School of Computer Science, Carnegie Mellon University, PA, USA~(e-mail: [schmerl; garlan]@cs.cmu.edu)}
        
\thanks{$^{3}$J. Cámara is with ITIS Software, Universidad de Málaga, Málaga, Spain~(e-mail: jcamara@uma.es)} 
        
\thanks{$^{4}$J. M. O'Kane is with Department of Computer Science and Engineering, Texas A\&M University, TX, USA (e-mail: jokane@tamu.edu)}
        
\thanks{$^{5}$E. C. Czaplinski, and K. A. Dzurilla are with Jet Propulsion Laboratory, California Institute of Technology, CA, USA (e-mail: [Ellen.C.Czaplinski; Katherine.A.Dzurilla]@jpl.nasa.gov)} 

\thanks{*Code and data are available at \url{https://github.com/softsys4ai/care}.}
}
      
% \thanks{*This work was supported by NASA (Award 80NSSC20K1720) and NSF~(Award 2107463). We thank Hari Nayar, Michael Dalal, Ashish Goel, Erica Tevere, Anna Boettcher, Anjan Chakrabarty, Ussama Naal, Carolyn Mercer, Issa Nesnas, Matt DeMinico, Md Shahriar Iqbal, and Jianhai Su for contributions to the \textsc{CaRE} framework and evaluations on the NASA testbeds: \url{https://nasa-raspberry-si.github.io/raspberry-si/}}% <-this % stops a space

\maketitle

\begin{abstract}
Robotic systems have subsystems with a combinatorially large configuration space and hundreds or thousands of possible software and hardware configuration options interacting non-trivially. The configurable parameters are set to target specific objectives, but they can cause functional faults when incorrectly configured. Finding the root cause of such faults is challenging due to the exponentially large configuration space and the dependencies between the robot's configuration settings and performance. This paper proposes \textsc{CaRE}---a method for diagnosing the root cause of functional faults through the lens of causality. \textsc{CaRE} abstracts the causal relationships between various configuration options and the robot’s performance objectives by learning a causal structure and estimating the causal effects of options on robot performance indicators. We demonstrate \textsc{CaRE}'s efficacy by finding the root cause of the observed functional faults and validating the diagnosed root cause by conducting experiments in both physical robots~(\textit{Husky} and \textit{Turtlebot~3}) and in simulation~(\textit{Gazebo}). Furthermore, we demonstrate that the causal models learned from robots in simulation (e.g., \textit{Husky} in \textit{Gazebo}) are transferable to physical robots across different platforms (e.g., \textit{Husky} and \textit{Turtlebot~3}).
\end{abstract}


%%%% arXiv version
\section{Introduction}
\IEEEPARstart{R}{obotic} systems are highly configurable, typically composed of multiple subsystems (e.g., localization, navigation), each of which has numerous configurable components (e.g., selecting different algorithms in the planner). Once an algorithm has been selected for a component, its associated parameters must be set to the appropriate values~(e.g., {\texTTT{use grid path = True}}). For instance, Arducopter is a robotic system with 622 configurable parameters, each with numerous valid values~\cite{taylor2016co}. To configure the ROS navigation stack for a particular robot, there are over 220 listed parameters~\cite{quigley2009ros}. The configuration space in such robotic systems is combinatorially large, with hundreds if not thousands of software and hardware configuration choices that interact non-trivially with one another. Indeed, incorrectly specified configuration options are one of the most common causes of system failure~\cite{HanEmpirical2016, iqbal2022unicorn, SiegmundPerformance2015,PEREIRA2021111044, 8451922}. 
\begin{figure}[!t]
 \vspace{-0.5em} 
  \centering
  \subfloat[]{\includegraphics[width=.36\columnwidth]{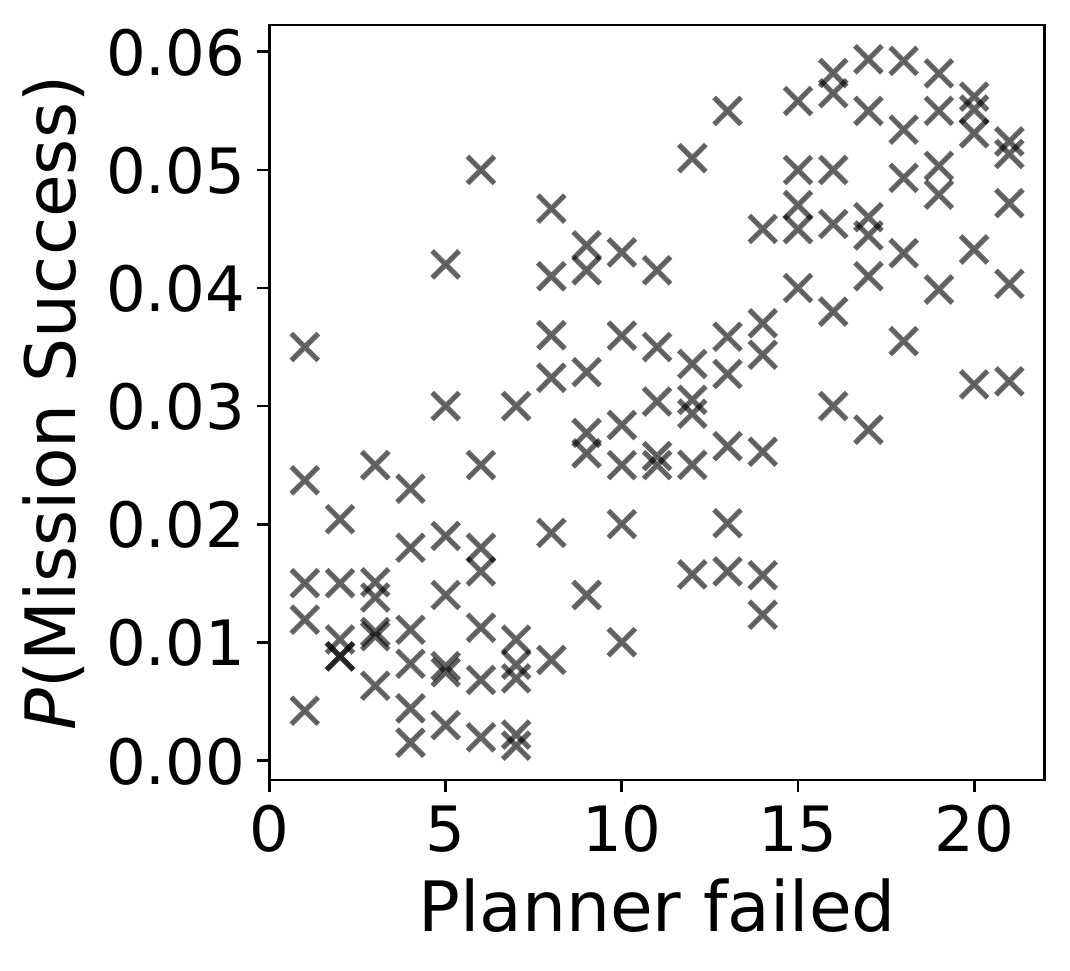}\label{fig:ml}}
  \hfill
  \subfloat[]{\includegraphics[width=0.36\columnwidth,]{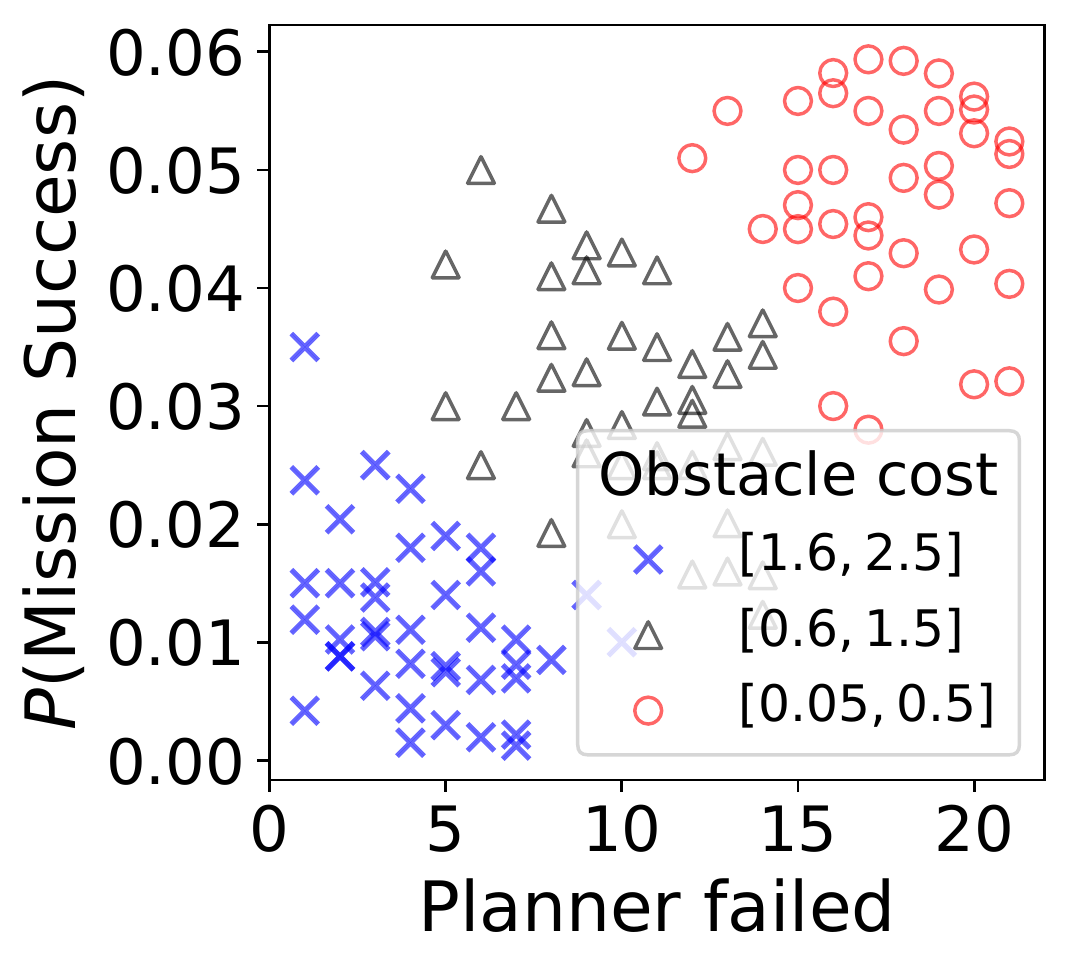}\label{fig:causal}}
  \hfill
  \subfloat[]{\includegraphics[width=0.27\columnwidth,]{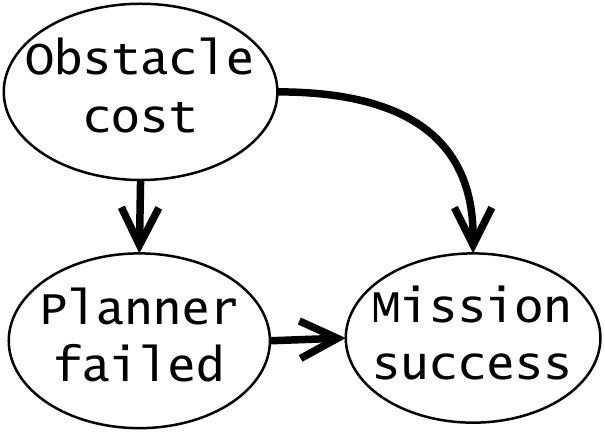}\label{fig:cm}}  
  \caption{\small{An example showing the effectiveness of causality in reasoning about the robot's behavior. (a) Observational data~(incorrectly) shows that an increase in the planner failure rate for producing a path leads to a higher probability of mission success; (b) incorporating {\texTTT{obstacle cost}} along the trajectory as a confounder correctly shows an increase in planner failure corresponding to a decrease in probability of mission success~(negative correlation); (c) the causal model correctly captures {\texTTT{obstacle cost}} as a common cause to explain the robot's behavior.}}
  \label{fig:intro}
 \vspace{-1.5em} 
\end{figure}

The configuration options in robotic systems directly impact mission objectives (e.g., navigating from one place to another), enabling trade-offs in the objective space~(e.g., the time that it takes to reach the target location(s) vs. the energy consumption for the task). The magnitude of the trade-off (even for the same configuration option) is dictated by its operating environment. The operating environment~($\mathpzc{E})$ may change at run-time beyond what was specified during design time $(\mathpzc{E}\rightarrow\mathpzc{E'})$, thereby potentially violating previously specified domain assumptions $\mathcal{C}, \mathpzc{E'} \not\models objective$, where $\mathcal{C}$ is a set of configuration options assigned to specified values before deployment. The configuration options impacting the objectives from the entire configuration space~$(\Tilde{C}\subset\mathcal{C}\rightarrow\Tilde{\mathcal{C}}')$ must be modified to satisfy the objectives $\Tilde{\mathcal{C}}', \mathpzc{E'} \models objective$. Unfortunately, configuring robotic systems to meet specified requirements is challenging and error-prone~\cite{timperley2022rosdiscover, swarm52020, kitz2021, EricLiu2019, viswa1211, torugobeck, automate}. Incorrect configuration~(called \textit{misconfiguration}) can cause buggy behavior, resulting in both \textit{functional} and/or \textit{non-functional faults}.\footnote{We define \textit{functional faults} as failures to accomplish the mission objective~(e.g., the robot could not reach the target location(s) specified in the mission specification). The \textit{non-functional faults} (interchangeably used as \textit{performance faults}) refer to severe performance degradation (e.g., the robot reached the target location(s); however, it consumed more energy, or it took more time than the specified performance goal in the mission specification).} Misconfigured parameters specified during design time can cause unexpected behavior at run time~\cite{mcarr2018, dneuhold, Werewolf86, AlexandrosNic}. In addition, the operating environment may change during a mission~\cite{liu2020mapper, merkt2019continuous, brasch2018semantic} and may require changing the configuration values on the fly~\cite{schmitt2019planning}. The aforementioned challenges make debugging robots a difficult task.

To handle the challenges in performance debugging and analysis, performance influence models~\cite{MartinComparison2021, 8919029,  chen2022performance, muhlbauer2020identifying} have received a significant amount of attention. Such models predict the performance behavior of systems by capturing the important options and interactions that influence the performance behavior. However, performance influence models built using predictive methods suffer from several shortcomings, including (i)~failing to capture changes in the performance distribution when deployed in unexpected environments~\cite{9345478}, (ii)~producing incorrect explanations as illustrated in Fig.~\ref{fig:causal}, (iii)~lack of transferability among common hardware platforms that use the same software stack~\cite{jamshidi2017transfer}, and (iv)~collecting the training data for predictive models from physical hardware is expensive and requires constant human supervision~\cite{gupta2018robot}. Traditional statistical debugging techniques~\cite{allamanis2018survey} based on correlational predicates, such as Cooperative Bug Isolation~(CBI), can be used to debug system faults. However, statistical debugging is hindered by the need for large-scale data and the inherent difficulty of pattern recognition in high-dimensional spaces~\cite{wong2016survey}, which can be challenging for robotic systems with non-linear interactions between variables.

To address this problem, we present an approach called \textsc{CaRE} (\underline{Ca}usal \underline{R}obotics D\underline{E}bugging) to diagnose the root causes of functional faults caused by misconfigurations in highly-configurable robotic systems through the lens of causality. \textsc{CaRE} works in three phases:  In Phase~\Romannum{1}, we first learn a causal model from observational data---dynamic traces measuring the performance objectives (e.g., energy, mission success, etc.) while the robot performs a mission under different configuration settings.
%To reduce the number of variables in the model, 
The causal model captures the causal relationships between configuration options and the robot's performance objectives. In Phase~\Romannum{2}, we use the causal graph to identify the \textit{causal paths}---paths that lead from configuration options to a performance objective. Next, in Phase~\Romannum{3}, we determine the configuration options with the highest causal effect on a performance objective by measuring each path's average causal effect to diagnose the functional faults' root causes. Our contributions are as follows:
\begin{itemize}
    \item We propose \textsc{CaRE} (\S\ref{sec:CARE}), a novel framework for finding the root causes of the configuration bugs in robotic systems.
    \item We evaluate \textsc{CaRE}, conducting a comprehensive empirical study (\S\ref{sec:exp}) in a controlled environment across multiple robotic platforms, including \textit{Husky} and \textit{Turtlebot~3} both in simulation and physical robots.
    \item We demonstrate the transferability of the causal models by learning the causal model in the \textit{Husky} simulator and reusing it in the \textit{Turtlebot~3} physical platform (\S\ref{sec:transferability}).
\end{itemize}
\section{Problem Description}
\subsection{Motivating scenarios} \label{sec:motivating scenario}
\subsubsection{Unmanned ground vehicles (UGV)}
To motivate the approach, we use the DARPA Subterranean Challenge~\cite{rouvcek2019darpa} to illustrate the following scenarios. This setting requires autonomous ground robots to work in adverse environments such as fog, debris, dripping water, or mud, and to navigate sloped, declining, and confined passageways. In this case, the mission objective is to stop the robot perpendicular to the position of a particular artifact and transmit its location to the control station.

\begin{figure}[!t]
 \vspace{-1em} 
  \centering
  \subfloat[Configuration bug]{\includegraphics[width=0.5\columnwidth,]{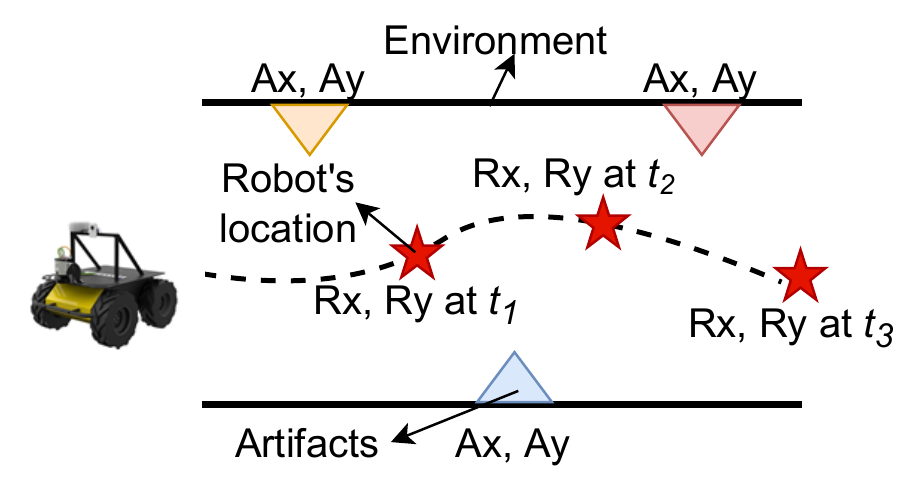}\label{fig:failedmission}}
  \hfill
  \subfloat[Change in environment]{\includegraphics[width=0.5\columnwidth,]{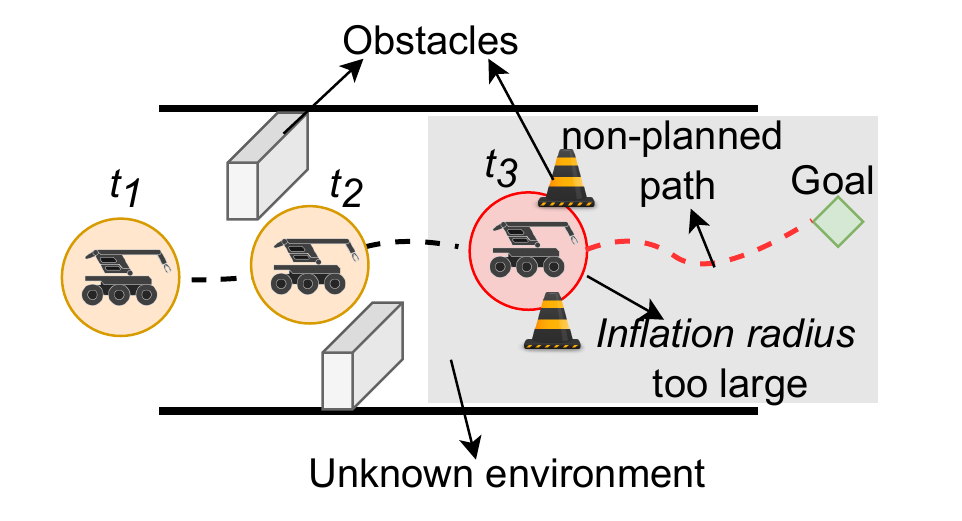}\label{fig:scenario2}}  
  \caption{\small{Different functional faults. (a) Delay in data transform results in a functional fault where the robot stops $0.5 m$ away from the target location and transmits incorrect artifact locations; (b) change in environment results in an indecisive robot that is stuck in place, where circles surrounding the robot represents the {\texTTT{inflation radius}}.}}
  \label{fig:scenario1}
   \vspace{-1.5em} 
\end{figure}

\paragraph{Functional fault due to configuration bug}  Fig.~\ref{fig:scenario1} shows a scenario where the robot stops $0.5$~m away from the target location and transmit incorrect artifact locations to the control station (as in Fig.~\ref{fig:scenario1}). A cause for this fault might be delay in data transformation~\cite{costmap2d}. For instance, the sensor transmits data at a rate of $1$~Hz and the robot travels at $0.5$~m/s. As a result, when the {\texTTT{costmap}} (which stores and updates information about obstacles in the environment using sensor data) receives data from the sensor, it is a second old and the robot has already traveled $0.5$~m away from that position. If the information of the {\texTTT{costmap}} is not immediate, it may cause the {\texTTT{global}} or {\texTTT{local planner}}~(components in robot navigation stack responsible for path planning) to make a wrong decision.

\paragraph{Functional fault due to change in environment}
Extending the previous scenario, suppose the obstacle locations are unknown to the robot. As shown in Fig.~\ref{fig:scenario2}, obstacles represent environment variables and circles surrounding the robot visualize a robot configuration parameter, {\texTTT{inflation radius}} (which specifies the object's maximum sensing distance). For example, setting {\texTTT{inflation radius}} $= x$ means that the robot should treat any pathways that remain $x$ meters or more away from obstacles. Fig.~\ref{fig:scenario2} shows a scenario where at $t_3$ the robot encounters unique obstacles that are too close together, violating the {\texTTT{inflation radius}}, defined prior to deployment, resulting in an indecisive robot that is stuck in place.

\paragraph{Incorrect reasoning about the robot's behavior} We perform a simple experiment for robot navigation, recording the number of failures in path planning {\texTTT{(planner failed)}} and probability of {\texTTT{mission success}}.
Fig.~\ref{fig:ml} shows the distribution of the $P${\texTTT{(mission success)}} with respect to {\texTTT{planner failed}}. We observe that an increase in {\texTTT{planner failed}} leads to a higher $P${\texTTT{(mission success)}}, which is counterintuitive. Such trend is typically captured by statistical reasoning in ML models. Incorporating {\texTTT{obstacle cost}} along the trajectory as a confounder~(Fig.~\ref{fig:causal}), correctly shows an increase in {\texTTT{planner failed}} corresponding to a decrease in the $P${\texTTT{(mission success)}} (negative correlation). The causal model (Fig.~\ref{fig:cm}) correctly captures {\texTTT{obstacle cost}} as a common cause to explain the correct relation  between the {\texTTT{planner failed}} and $P${\texTTT{(mission success)}}. The arrows denote the assumed direction of causation, whereas the absence of an arrow shows the absence of direct causal influence between variables such as configuration options, and performance objectives.

\begin{figure}[!t]
 \vspace{-1em} 
  \centering
  \subfloat[A failed excavation task]{\includegraphics[width=.49\columnwidth]{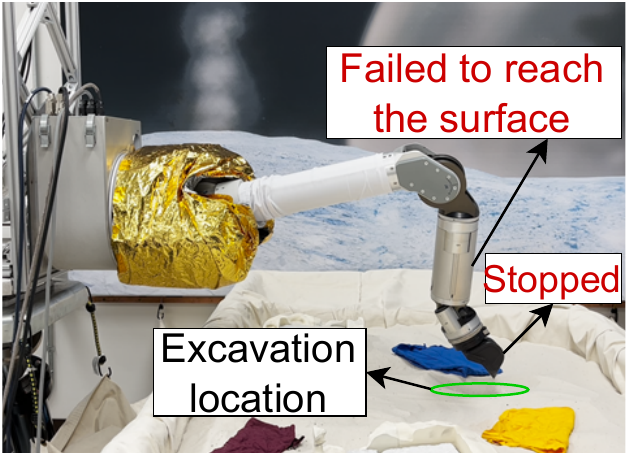}\label{fig:olwat_failed}}
  \hfill
  \subfloat[A successful excavation task]{\includegraphics[width=0.5\columnwidth,]{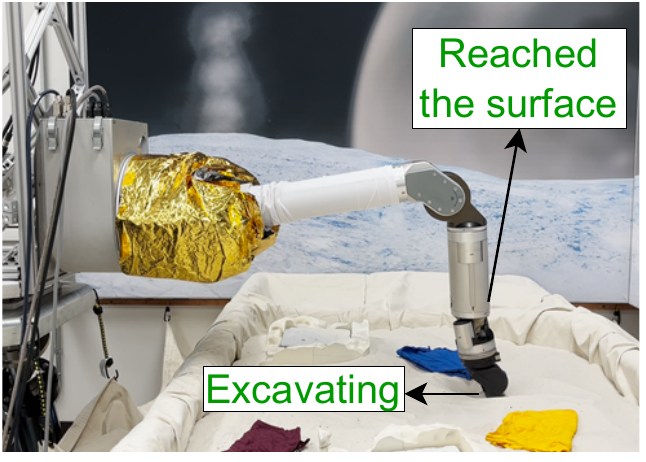}\label{fig:owlat_pass}}
  \caption{\small{OWLAT hardware platform performing excavation task; (a) the excavation task failed due to the robotic arm was unable to find the surface, (b) the desire excavation operation without the fault.}
  \label{fig:owlatscenario}}
   \vspace{0.5em} 
\end{figure}

\subsubsection{Robotic exploration of Ocean Worlds}
We use Europa Lander Mission~\cite{europa} as a test case to illustrate the following scenarios. The lander mission is expected to last at least 20 days, during which the lander is expected to complete a total of at least 5 sample acquisitions. Due to high-radiation levels in the environment, resources and data would be tightly constrained, and high communication overhead requires the lander operations highly automated. We use Ocean World Lander Autonomy Testbed (OWLAT)~\cite{owlat} testbed platform as an example of a planetary robotic system. The testbed comprises a base, an instrument arm, and perception sensors (e.g., camera). The description of the OWLAT platforms are as follows:
\begin{itemize}
    \item OWALT Hardware Platform (owlat-physical): It provides the hardware interfaces and low-level software infrastructure to allow various autonomy solutions to command typical lander operations and receive telemetry and performance feedback through a ROS-based software interface in a plug-and-play manner.
    \item OWLAT Control Software Simulation (owlat-sim): This software provides a partial simulation of the OWLAT hardware platform via Dynamics and Real-Time Simulation (DARTS). The simulation includes the robotic arm commands via ROS actions.
\end{itemize}

\noindent The OWLAT platform includes different action commands, each of which has numerous configurable inputs. A subset of the different action commands and their respective inputs are shown in Table~\ref{tab:owlat_config}. To perform a task (say, excavate at a location $(x, y, z)$), a series of action commands must be executed in a sequence. Once the sequence of the action commands has been decided, its associate inputs must be set to the appropriate values.

\setlength{\textfloatsep}{0pt}% Remove \textfloatsep
\begin{table}[!t]
\centering
\caption{A subset of the configuration space of the OWLAT platform.}
\label{tab:owlat_config}
\scalebox{0.9}{
\begin{tabular}{cl} 
\toprule
\textbf{Action Command} & \textbf{Input} \\ 
\hline
\multirow{2}{*}{ARM MOVE JOINT} & {\texTTT{relative:}}~{\texTTT{bool}} \\
 & {\texTTT{angles:~float[7]}} \\ 
\hline
\multirow{5}{*}{\begin{tabular}[c]{@{}c@{}}ARM MOVE JOINTS\\GUARDED\end{tabular}} & {\texTTT{relative:~bool}} \\
 & {\texTTT{angles:~float[7]}} \\
 & {\texTTT{retracting:~bool}} \\
 & {\texTTT{force threshold:~float}} \\
 & {\texTTT{torque threshold:~float}} \\ 
\hline
\multirow{7}{*}{\begin{tabular}[c]{@{}c@{}}ARM MOVE CARTESIAN\\GUARDED\end{tabular}} & {\texTTT{frame:~int}} \\
 & {\texTTT{relative:~bool}} \\
 & {\texTTT{position:~[x, y, z]}} \\
 & {\texTTT{orientation:~[qw, qx, qy, qz]}} \\
 & {\texTTT{retracting:~bool}} \\
 & {\texTTT{force threshold:~float}} \\
 & {\texTTT{torque threshold:~float}} \\
\bottomrule
\end{tabular}}
\end{table}
\setlength{\textfloatsep}{0pt}% Remove \textfloatsep

\paragraph{Functional fault due to uncertainties existing in domain assumptions} Fig.~\ref{fig:olwat_failed} such a scenario where the robotic arm encounters an {\texTTT{arm goal error}} fault (which represents the arm failed to find the surface), during the excavation task, as a result the arm stopped in place and resulted in the excavation failure. The system usually encounters such functional fault due to uncertainties existing in domain assumptions such as incorrectly simplified environment model (e.g., incorrectly predicting the hardness of the surface), resulting in the autonomy assigns a wrong number to a constant variable (e.g., incorrect {\texTTT{force threshold}} value).

\subsubsection{Challenges}
The above scenarios illustrate that it can be challenging to identify the underlying causes of functional faults due to the following reasons:
\setlength{\textfloatsep}{0pt}% Remove \textfloatsep
\begin{itemize}
    \item A typical debugging approach to find the root causes of such functional faults might be trial-and-error. However, this process requires non-trivial human effort due to the large configuration space.
    \item Another common strategy for performance debugging is to build performance influence models, such as regression models, by understanding the influence of the individual configuration options and their interactions by inferring the correlations between configuration options and performance objectives. However, performance influence models are unable to capture changes in the performance distribution when deployed in a different environment, as a result these models are typically non-transferable in the unseen environment.
    \item Collecting the training data from the physical hardware is expensive and requires a human in the loop to avoid any damage to the hardware.
    \item Robotic systems are often deployed in adverse environments where the environment may change during a mission (as in Fig.~\ref{fig:scenario2}) and may require changing the configuration values on the fly.
\end{itemize}

\subsection{Causal reasoning for robotics}\label{sec:problem formulation}
We formulate the problem of finding root causes for functional faults in robotic systems using an abstraction of causal model utilizing \textit{Directed Acyclic Graphs} (DAGs)~\cite{pearl1998graphical}.
The causal model encodes performance variables, functional nodes~(which defines functional dependencies between performance variables such as how variations in one or multiple variables determine variations in other variables), causal links that interconnect performance nodes with each other via functional nodes, and constraints to define assumptions we require in performance modeling (e.g., the configuration options cannot be the child node of performance objectives). Table~\ref{tab:notations} summarizes the notations used in the following sections.
\setlength{\textfloatsep}{10pt}% Remove \textfloatsep
\begin{table} [!t]
\vspace{0.5em}
\centering
\caption{Summary of notations used in \S\ref{sec:problem formulation} and \S\ref{sec:CARE}}
\label{tab:notations}
\resizebox{\linewidth}{!}{%
\begin{tabular}{cl} 
\toprule
\textbf{Notation} & \multicolumn{1}{c}{\textbf{Description}} \\ 
\hline
$\mathcal{X}$ & Manipulable variables (e.g., configuration options) \\
$\mathcal{S}$ & Non-manipulable variables (e.g., performance metrics) \\
$\mathcal{Y}$ & Performance objectives \\
$D$ & Observational data, $\{\mathcal{X}, \mathcal{S}, \mathcal{Y}\}$ \\
$\mathcal{G}$ & Dense graph \\
$\mathcal{G}_\partial$ & Partial ancestral graph \\
$\mathcal{V}$ & Vertex set, $\{v_0, \dots, v_n\}$ \\
$\mathbb{P}$ & Causal path, $\langle P_0, \dots, P_n \rangle$  \\
$\mathcal{D}$ & Directed edge, $v_i\ \edgeone\ v_j$ \\
$\mathcal{B}$ & Bi-directed edge, $v_i\ \edgetwo\ v_j$ \\
$\partial$ & Partial edge, $\{\partial_1 = v_i\ \edgethree\ v_j,\ \partial_2 =v_i\ \edgefour\ v_j\}$ \\
$S_c$ & Structural constraints (e.g., no $\mathcal{V}[objective]\ \edgeone\ \mathcal{V}[option]$~ \\
$\mathcal{Z}$ & Unmeasured confounder between $v_i$ and $v_j$ \\
\bottomrule
\end{tabular}
}
%\vspace{-1em}
\end{table}

Given a robotic system that intermittently encounters functional faults, our goal is to find the root causes of the functional faults by querying a causal model learned from the observational data. We start by formalizing the problem of finding the causal directions from configuration options to performance objectives that indicate a functional fault. This problem can further be subdivided into two parts: (a)~learning---discovery of the true causal relationship between nodes, and (b)~inference---identification of the root causes for a functional fault using the learned causal model. Let us consider a configurable robotic system~$\mathcal{A}$ which has a set of manipulable (or configurable) variables~$\mathcal{X}$  that can be intervened upon, a set of non-manipulable variables~$\mathcal{S}$ (non-functional properties of the system such as metrics that evaluate the performance) that can not be intervened, and a set of performance objectives~$\mathcal{Y}$. We define the causal graph discovery problem formally as follows:

\vskip 0.2em
\noindent \fbox{\begin{minipage}{24.5em}
\noindent \textbf{Problem 2.1} (\textit{Learning}). 
Given observational data $D$, recover the causal graph $\mathcal{C_G}$
%characterized 
%by the true causal graph $\mathcal{C_G}$ 
that encodes the dependency structure between $\mathcal{X}, \mathcal{S}$, and $\mathcal{Y}$ of $\mathcal{V}$ such that the following structural constraints are satisfied.
\begin{align*}
    v_i \nleftrightline v_j \ \forall v_i \in \mathcal{X} \subset \mathcal{V} , \ \forall v_j \in \mathcal{Y} \subset \{\mathcal{V} \setminus \mathcal{X} \setminus \mathcal{S}\}
\end{align*}
\end{minipage}}
\vskip 0.5em

\noindent The second part of the problem is to find the root cause of functional fault using the learned causal model. We formulate the inference problem to estimate the average causal effect of the configuration option on the performance objectives as:

\noindent \fbox{\begin{minipage}{24.5em}
\noindent \textbf{Problem 2.2} (\textit{Inference}). 
Given the causal graph $\mathcal{C_G}$, determine the configuration option in $\mathcal{X}$ which is the root cause for the observed functional fault characterized by performance objectives $\mathcal{Y}$ as follows:
%\vspace{-0.2em}
\begin{align*}
    \{v_i^*\}  = \argmax_{v_i} ACE(v_i,v_j^*)
    %\mathbb{E}[v_j^*|do(V=v_i)]
\end{align*}
where $\{v_i^*\} \subset \mathcal{X}$ is the set of root causes (configuration options), $\{v_j^*\} \subset \mathcal{Y}$ are the performance objectives characterizing the functional fault, and $ACE$ represents the average causal effect---the average difference between potential outcomes under different treatments\cite{pearl2009causality}.
\end{minipage}}
\section{\textsc{CaRE}: \underline{Ca}usal \underline{R}obotics D\underline{E}bugging} \label{sec:CARE}
\begin{figure}[!t]
\vspace{-0.2em}
\begin{center}
\centerline{\includegraphics[width=0.95\columnwidth]{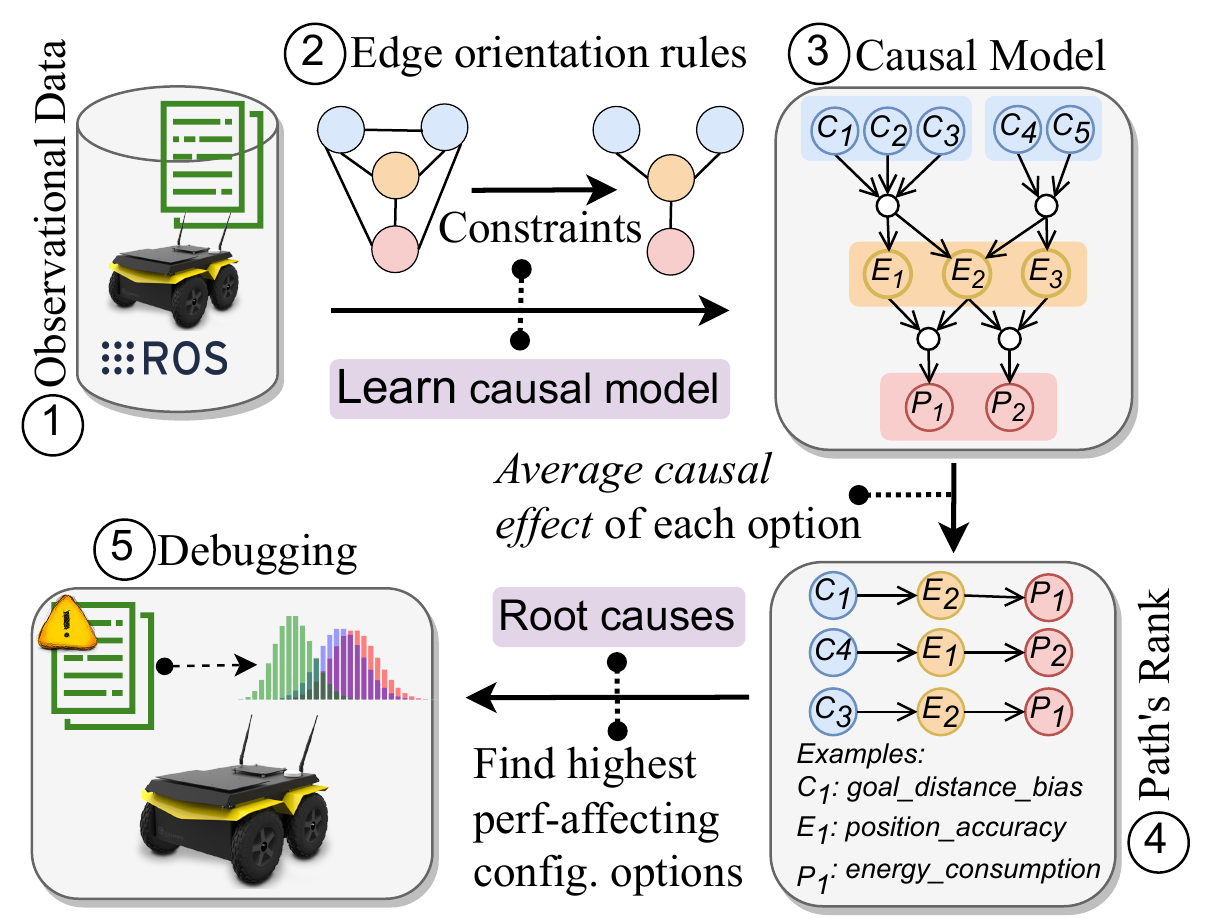}}
\caption{Overview of \textsc{CaRE}}
\label{fig:care_overview}
\end{center}
 \vspace{-2em} 
\end{figure}

We propose a novel approach, called \textsc{CaRE}, to find and reason about the intricate relations between configuration options and their effect on the performance objectives in highly configurable robotic systems. \textsc{CaRE} works in three phases: (i) The observational data is generated by measuring the performance metrics and performance objectives under different configuration settings (see \circled{1} in Fig.~\ref{fig:care_overview}) to construct the graphical causal model (see \circled{3} in Fig.~\ref{fig:care_overview}) enforcing the structural constraints (see \circled{2} in Fig.~\ref{fig:care_overview}). (ii) The causal model is used to determine the paths that lead from configuration options to the performance objectives (see \circled{4} in Fig.~\ref{fig:care_overview}). (iii) The configuration options that has the highest causal effect on the performance objective was determined by measuring the average causal effect of each path to fix or debug the configuration issues (see \circled{5} in Fig.~\ref{fig:care_overview}).

\subsection{Learning the causal model} \label{sec:cpm}
We design a three-layer structure causal model defining three variable types: (i)~software-level configuration options associated with different algorithms (e.g., {\texTTT{goal distance bias}}~\cite{navcore}), and hardware-level options (e.g., {\texTTT{sensor frequency}}), (ii)~intermediate performance variables~(non-manipulable variables) that map the influence of the configuration options to the performance objectives~(e.g., {\texTTT{position accuracy}}), and (iii)~end-to-end performance objectives (e.g., energy). We classify the performance variables as non-manipulable and manipulable variables to reduce the number of variables that require intervention. Note that the level of debugging can vary~\cite{jamshidi2017transfer}, and the abstraction level of the variables in the causal model depends on the debugger and can go all the way down, even to the hardware level~\cite{iqbal2022unicorn}. To build the three-layer structure, we define two specific constraints over causal models: (i)~variables that can be manipulated (e.g., using prior experience, the user may want to restrict the variables that do not have a significant impact on performance objectives); (ii)~structural constraints~(e.g., configuration options do not cause other options). Such constraints enable incorporating domain knowledge that facilitates learning with low sample sizes.

Several methods are proposed to extract the causal graphical model from data in the literature. These belong into two categories: constraint-based techniques and score-based techniques. We specifically use \textit{Fast Causal Inference}~(hereafter, FCI)~\cite{spirtes2000causation}, a constraint-based technique for identifying the causal model guiding robot performance. We select FCI because it identifies the unobserved confounders (latent common causes that have not been, or cannot be, measured); and it can handle various data types (e.g., nominal, ordinal, and categorical) given a valid conditional independence test. When the FCI algorithm is applied to observational data, a \textit{Partial Ancestral Graph} (PAG)~\cite{colombo2012learning, glymour2019review}, representing a causal structure in the presence of latent variables, is produced. Each edge in the PAG denotes the ancestral connections between the vertices. For a comprehensive theoretical foundation on these ideas, we refer the reader to~\cite{colombo2012learning, colombo2014order, pearl2016causal}. To discover the true causal relationship between two variables, the causal graph must be fully resolved~\cite{iqbal2022unicorn} such that there are no $v_i\ \edgethree\ v_j$ ($v_i$ causes $v_j$, or there are unmeasured confounders that cause both $v_i$ and $v_j$), and $v_i\ \edgefour\ v_j$ ($v_i$ causes $v_j$, or $v_j$ causes $v_i$, or there are unmeasured confounders that cause both $v_i$ and $v_j$) edges. We define the partial edge resolving problem formally as follows:
\vskip 0.1em
\noindent \fbox{\begin{minipage}{24.5em}
\noindent \textbf{Problem 3.1} (\textit{Resolve Partially Directed Edges}.) Given a causal partial ancestral graph~\cite{glymour2019review} $\mathcal{\mathcal{G}_\partial} = (\mathcal{V}, \partial)$, the partial edge resolving problem involves replacing each partial edge $\partial$ with a directed edge $\mathcal{D}$ or a bi-directed edge~$\mathcal{B}$ based on some threshold $\theta$.
\end{minipage}}
\vspace{0.2em}

We use Algorithm~\ref{alg:cpm} for learning the causal model~(CM). First, we build a dense graph~$\mathcal{G}$ by connecting all pairs of configuration options, performance metrics, and performance objective with an undirected edge. The performance metrics (non-manipulable variables), maps the influence of the configuration options to performance objectives. Unlike configuration options, the intermediate layer's variables can not be modified. However, they can be observed and measured to understand how the causal-effect of changing configurations propagates to mission success, energy consumption. The skeleton of the causal model is recovered by enforcing the structural constraints (e.g., no connections between configuration options, as in line 3 of Algorithm~\ref{alg:cpm}). Next, we evaluate the independence of all pairs of variables conditioned on all remaining variables using Fisher’s exact test~\cite{connelly2016fisher}. A partial ancestral graph (PAG) is generated, orienting the undirected edges by employing the edge orientation rules~\cite{spirtes2000causation,glymour2019review,colombo2012learning} (line 4 of Algorithm~\ref{alg:cpm}). The obtained PAG must be fully resolved (no $\partial$ between two vertices) to discover the true causal relationships. We resolve the FCI-generated PAG by evaluating if an unmeasured confounder $(\mathcal{Z})$ is present between two partially oriented nodes ($v_i, v_j$). Employing the information-theoretic approach based on entropy~\cite{Kocaoglu2020} produces a joint distribution $q(v_i, v_j, \mathcal{Z})$. We compute the entropy $H(\mathcal{Z})$ of $\mathcal{Z}$. Comparing the $H(\mathcal{Z})$ with $\theta_r$ (entropy threshold, $\theta_r = 0.8 \min\{H(v_i), H(v_j)\}$), we determine $\forall\ \mathbb{P}$ if $\exists \mathcal{Z} \in \partial$, as shown in lines 6-17 of Algorithm~\ref{alg:cpm}, where $E$ and $\hat{E}$ are the extrinsic variables responsible for system noise ($v_i \ind E$, $v_j \ind \hat{E}$). The final causal model is an \textit{acyclic directed mixed graph} (ADMG)~\cite{richardson2002ancestral}.
\setlength{\textfloatsep}{0pt}% Remove \textfloatsep
\begin{algorithm} [!t]
\DontPrintSemicolon
\SetKwFor{ForEach}{for each}{do}{end}
\SetKwRepeat{Do}{do}{while}
\caption{CM$(\textrm{data}, \mathcal{V}, \mathcal{G})$}\label{alg:cpm}
\KwIn{data, $\mathcal{G}$, Vertex set $\mathcal{V}$~$\gets$ \{$(\mathcal{X}_1, \dots, \mathcal{X}_n)$, $(\mathcal{S}_1, \dots, \mathcal{Y}_m)$, $(\mathcal{Y}_1, \dots, \mathcal{Y}_p)\}$}
\KwOut{Set of $\mathcal{D}$ and $\mathcal{B}$ to build the ADMG}
        $\mathcal{D} \gets \emptyset$, $\mathcal{B} \gets \emptyset$  \\
        \While{$\mathcal{V} \in \mathcal{G}$}{
        $S_c \gets$ Apply structural constraints on $\mathcal{G}$\\
        $\mathcal{G}_\partial = \text{FCI}(data,\ \text{Fisher z-test},\ S_c)$ \\
        \ForEach{$\partial \in \mathcal{G}_{\partial}$}{
            Compute entropies $H(v_i), H(v_j), H(\mathcal{Z})$ \\ 
            $\theta_r = 0.8 \min\{H(v_i), H(v_j)\}$ \\
            \eIf{$H(\mathcal{Z}) < \theta_r$}{
                Replace $v_i$\edgethree $v_j$ with $v_i$\edgetwo $v_j$ in $\mathcal{B}$}
            {$v_j = f(v_i,E)\ \text{where}\ [E \ind v_i]$ \\
                  $v_i = g(v_j,\hat{E})\ \text{where}\ [\hat{E} \ind v_j]$  \\  
                  Compute the entropies $H(E)$ and $H(\hat{E})$\\
                  {\If{$H(E)<H(\hat{E})$}{
                        %$\mathcal{D} \gets$ $(v_i\ \edgefour\ v_j\ \KwTo\ v_i\ \edgeone\ v_j)$}}
                        Replace $v_i\ \edgefour\ v_j$ with $v_i\ \edgeone\ v_j$ in $\mathcal{D}$}}
                    % \Else {$\mathcal{D} \gets$ $(v_i$ \edgefour $v_j$ \KwTo $v_j$ \edgeone $v_i)$}   
                    \Else {Replace $v_i\ \edgefour\ v_j$ with $v_j\ \edgeone\ v_i$ in $\mathcal{D}$}
        }}%}
\KwRet{$\mathcal{D}, \mathcal{B}$}
}
\end{algorithm}

\subsection{Causal effect estimation}
\label{lb:Causaleffectestimation}
To determine the root cause of a functional fault from  the causal graph, we need to extract the paths from $\mathcal{C_G}$ (referred to as \textit{causal paths}). A causal path is a directed path originating from $\mathcal{X}$ (e.g., configuration options) to a subsequent non-functional property $\mathcal{S}$ (e.g., performance metrics) and terminating at $\mathcal{Y}$ (e.g., performance objectives). Our goal is to find an ordered subset of $\mathbb{P}$ that defines the causal path from the root cause of the functional fault (a manipulable variable that cause the functional fault) to the performance objective indicating the functional fault (say $x_i$ causes a functional fault $F$ through a subsequent node $s_i$ in the path, assuming $(\exists x_i \in \mathcal{X}) \wedge (\exists s_i \in \mathcal{S}$); e.g., $x_i\ \edgeone\ s_i\edgeone\ \mathcal{Y}_F$). We define the causal path discovery problem as follows:
\begin{figure}[!ht]
\vspace{-1em}
\noindent \fbox{\begin{minipage}{24.5em}
\noindent \textbf{Problem 3.2} (\textit{Causal Path Discovery}). Given a causal graph $C_\mathcal{G}=(\mathcal{V}, \mathcal{D}, \mathcal{B})$ that encodes the dependency structure between $\mathcal{X, S}$ and $\mathcal{Y}$, and a performance objective $\mathcal{Y}_F \in \mathcal{V}$ indicating a specific functional fault, the causal path discovery problem seeks a path $\mathbb{P}=\langle v_0, v_1,\dots,v_n \rangle$ such that the following conditions hold:
\begin{itemize}[•]
    \item \small $v_o$ is the root cause of the functional fault and $v_n = \mathcal{Y}_F$.
    \item \small $\forall\ 0 \leq i \leq n,\ v_i \in \mathcal{V}$ and $\forall\ 0 \leq i \leq n,\ (v_i, v_{i+1}) \in (\mathcal{D} \vee\mathcal{B})$.
    \item \small $\forall\ 0 \leq i \leq j \leq n$, $v_i$ is a counterfactual cause of $v_j$.
    \item \small $\left|\mathbb{P}\right|$ is maximized.
\end{itemize}
\end{minipage}}
\vspace{-1em}
\end{figure}

We extract the causal paths and measure the average causal effect of the extracted causal paths on the performance objectives $(\mathcal{Y})$, and  rank the paths from highest to lowest using Algorithm~\ref{alg:CPWE}: Causal Paths With Effect (CPWE). CPWE simplifies the complicated causal graph using path extraction and ranking to a small number of useful causal paths to determine the configurations that have the highest influence on the performance objectives. A causal path is a directed path orientating from configuration options to performance metric and terminating at a performance objective. Causal paths are discovered by backtracking from the nodes corresponding to each of the performance objective until we reach a node with no parents. The discovered paths are then ranked by measuring the causal effect of a node's value change (say $V_1$) on its subsequent node $V_2$ in the path. We express this using the \textit{do-calculus}~\cite{pearl2009causality} notation: $\mathbb{E}\left[V_2~|~\text{do}\left(V_1=x\right)\right]$ that represents the expected value of $V_2$ if we set the value of the node $V_1$ to $x$. The \textit{average causal effect} (ACE) of $V_1$\edgeone $V_2$ was calculated across all acceptable $V_1$ values as follows:
\begin{equation}
    \resizebox{0.9\hsize}{!}{$\mathrm{ACE}\left(V_2, V_1\right) = \frac{1}{N} \sum_{\forall x, y\in V_1}\mathbb{E}\left[V_2~|~\text{do}\left(V_1=y\right)\right]~-~\mathbb{E}\left[V_2~|~\text{do}\left(V_1=x\right)\right]$},
    \label{eq:ace}
\end{equation}
where, $N$ is the total number of acceptable values of $V_1$. $\mathrm{ACE}\left(V_2, V_1\right)$ will be larger if $V-1$ yield a larger change in $V_2$. We calculate the ACE for the entire causal path extending Equation~\ref{eq:ace} as follows:
\begin{equation}
    \mathbb{P}_{ACE} = \frac{1}{K} \sum \mathrm{ACE}(v_j, v_i)
    \label{equ:rank}
\end{equation}
The configuration options that are found on paths with larger $\mathbb{P}_{ACE}$ are likely to have a higher causal effect on the corresponding performance objective. The top $K$ paths with the largest $\mathbb{P}_{ACE}$ values were selected for each of the performance objective.
\section{Experiments and Results} \label{sec:exp}
Using the \textit{Husky} and \textit{Turtlebot~3} platforms as case study systems, we answer the following research questions (RQ):
\begin{itemize}
    \item RQ1 (Accuracy): To what extent are the root causes determined by \textsc{CaRE} the true root causes of the observed functional faults?
    \item RQ2 (Transferability): To what extent can \textsc{CaRE} accurately detect misconfigurations when deployed in a different platform?
\end{itemize}

\setlength{\textfloatsep}{0pt}% Remove \textfloatsep
\begin{algorithm}[!t]
\DontPrintSemicolon
\SetKwFor{ForEach}{for each}{do}{end}
\SetKwRepeat{Do}{do}{while}
\caption{CPWE$(\mathcal{V}, \mathcal{D}, \mathcal{B})$}\label{alg:CPWE}
\KwIn{data, $\mathcal{V}$, $\mathcal{D}$, $\mathcal{B}$}
\KwOut{Rank of the causal paths}
        $\mathbb{P}\ \gets\ \emptyset$, $K\ \gets\ \emptyset$ \\
        ADMG $\gets\ \{\mathcal{V}, \mathcal{D}, \mathcal{B}\}$ \\
        \While{$\mathcal{V}[\mathcal{Y}] \in $ ADMG}{
            $\mathbb{P} \gets$ All causal paths from an $\mathcal{V}[\mathcal{Y}]$ node \\

            \For{$i \gets P_1\ \KwTo\ P_n$}{
                Compute $\mathbb{P}_{ACE}$ usign Equation~\ref{equ:rank} \\
                $K \gets \textsc{SortDescending}(\mathbb{P}_{ACE})$  
            }%}
        
\KwRet{$K$}
}
\end{algorithm}

\subsection{Experimental setup} \label{sec:exp_reval}
\begin{figure}[!t]
    \vspace{-1em}
  \centering
  \subfloat[Simulated environment]{\vstretch{0.9}{\includegraphics[width=.34\columnwidth, valign=t]{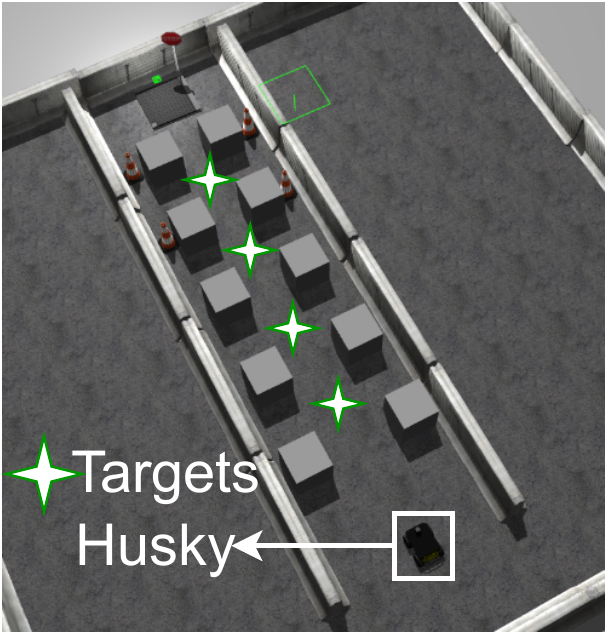}\label{fig:env}}}
  \hfill
  \subfloat[Real environment]{\includegraphics[width=0.64\columnwidth,valign=t]{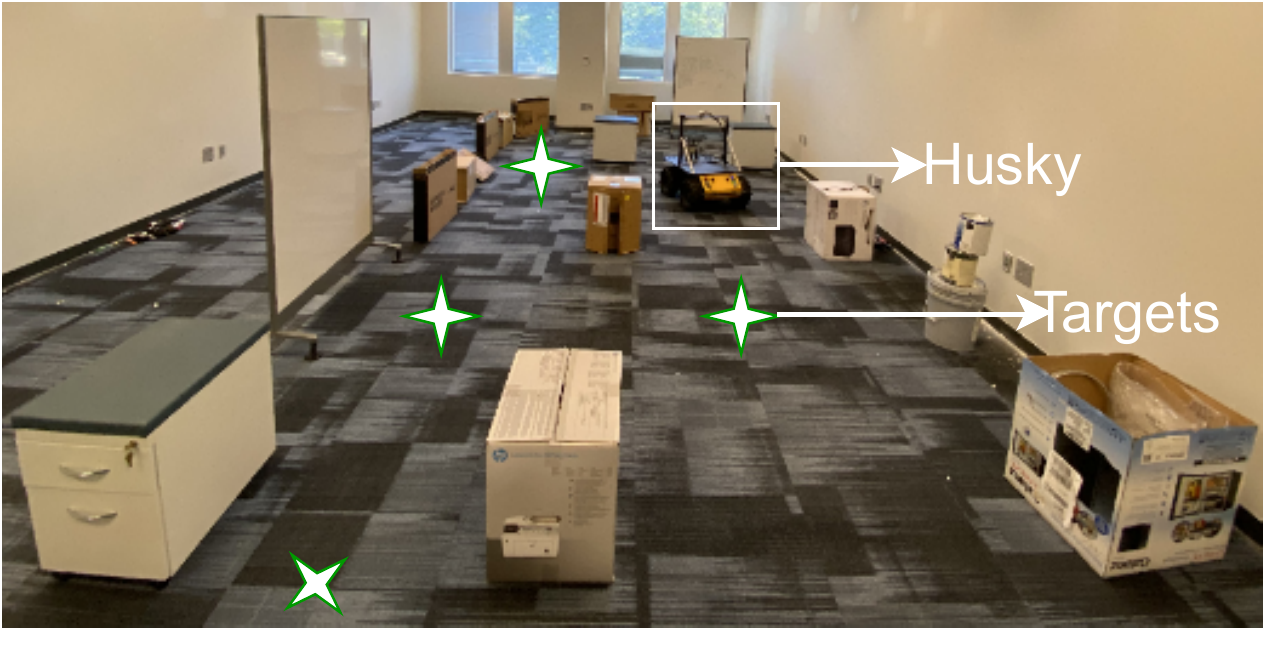}\label{fig:env_real}}
  \caption{\small{Experimental environments, (a) simulated in \textit{Gazebo}, (b) a real environment located at the University of South Carolina.}}
  \label{fig:exp_env}
  \vspace{0.5em} 
\end{figure}

We simulate \textit{Husky} in \textit{Gazebo} to collect the observational data  by measuring the performance metrics (e.g., traveled distance) and performance objectives (e.g., energy consumption) under different configuration settings (\S\ref{sec:config_options}) to train the causal model. Note that we use simulator data to evaluate the transferability of the causal model to the physical robots, but \textsc{CaRE} also works with data from physical robots. We deployed the robot in a controlled indoor environment and directed the robot to autonomously navigate to the five target locations~(Fig.~\ref{fig:exp_env}) using \textit{Husky} {\texTTT{move\_base\_mapless.launch}}. The robot was expected to encounter obstacles and narrow passageways, where the locations of the obstacles were unknown before deployment. The mission was considered successful if the \textit{Husky} robot reached each of the five target locations. We used Euclidean distance between the commanded and measured positions as a threshold to determine if a target was reached. Using random sampling, we generated the values for the configurable parameters (\S\ref{sec:sup-mat}) and recorded the performance metrics (the intermediate layer of the causal model that map the influence of the configuration options to the performance objective) for different values of the configurable parameters. We used the navigation task as a test case and defined the following performance metrics for the ROS Navigation Stack~\cite{navcore}:
\begin{enumerate}
    \item \textit{Traveled distance (TD)}: Traveled distance from start to destination.
    \item \textit{Robustness in narrow space (RNS)}: We define $\mathrm{narrow\ space} = \mathrm{Robot_{footprint}} + \mathrm{Footprint_{padding}}$, and $\mathrm{RNS}=\frac{1}{N_{s}}\sum_{i=1}^{N_{s}} \mathrm{Passed}_{N_{s}}$, where $N_{s}$ is the total number of narrow spaces in the known environment, and $\mathrm{Passed}_{N_{s}}$ is the narrow spaces that the robot successfully passed.
   \item \textit{Mission time}: Total time (\textit{minute}) to complete a mission.
    \item \textit{Recovery executed (RE)}: Number of {\texTTT{rotate recovery}} and {\texTTT{clear costmap recovery}} executed per mission.
    \item \textit{Replanning path (RP)}: Number of replanning paths performed by the planner during a mission.
    \item \textit{Error rotating to the goal (ERG)}: Number of errors when rotating to a goal per mission execution. If the robot reaches the goal and stops, we check if there is a potential collision while rotating. 
\end{enumerate}

\begin{figure}[!t]
\begin{center}
\centerline{\includegraphics[width=0.99\columnwidth]{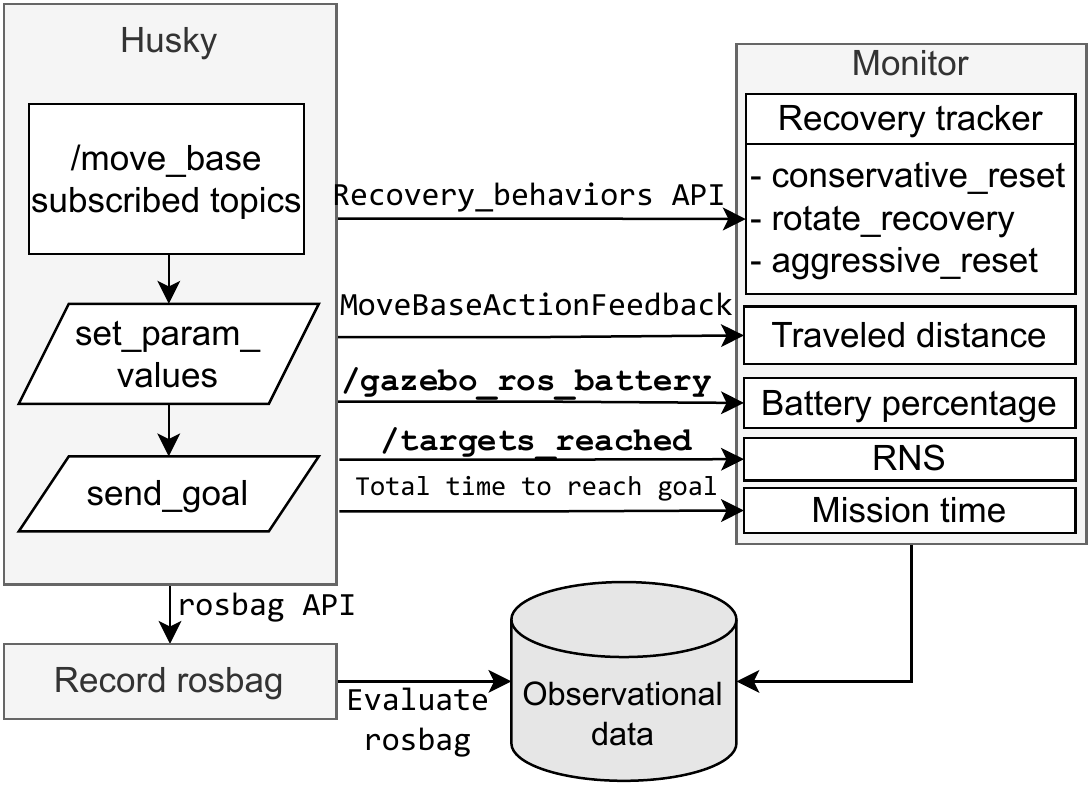}}
\caption{\small{\textit{Reval} interface for observational data collection.}}
\label{fig:data_collect}
\end{center}
\vspace{-0.5em} 
\end{figure}

\noindent Additionally, we integrate the Gazebo battery plugin~\cite{batteryplugin} to the \textit{Husky} simulator to measure the energy consumption. To collect the observational data, we developed \textit{Reval}\footnote{\url{https://github.com/softsys4ai/Reval.git}}--- a tool to evaluate ROS-based robotic systems. The observational data was collected while the \textit{Husky} performs a mission using the interface as shown in Fig.~\ref{fig:data_collect}. Additional details about experiments are provided in \S\ref{sec:sup-mat}.

\subsection{RQ1: Accuracy} \label{sec:vefification}
We answer RQ1 through an empirical study. Recall that our overall goal is to determine the parameters that influence the performance objective most. We analyze whether changing the value of a configuration option noticeably affects the performance distribution (options that have a stronger influence are likely to have high variability). In particular, we perform several experiments to validate the root causes determined by \textsc{CaRE} for both \textit{Husky} in simulation and in physical robot by comparing the \textit{variance}~($\sigma^2$) of the performance objectives (e.g., energy, mission success) and performance metrics (e.g., RNS, traveled distance) for different configuration options. 

\begin{figure}[!t]
  \centering
  \subfloat[A partial causal model for ROS navigation stack
discovered in our experiments using the \textit{Husky} simulator.]{\includegraphics[width=1\columnwidth]{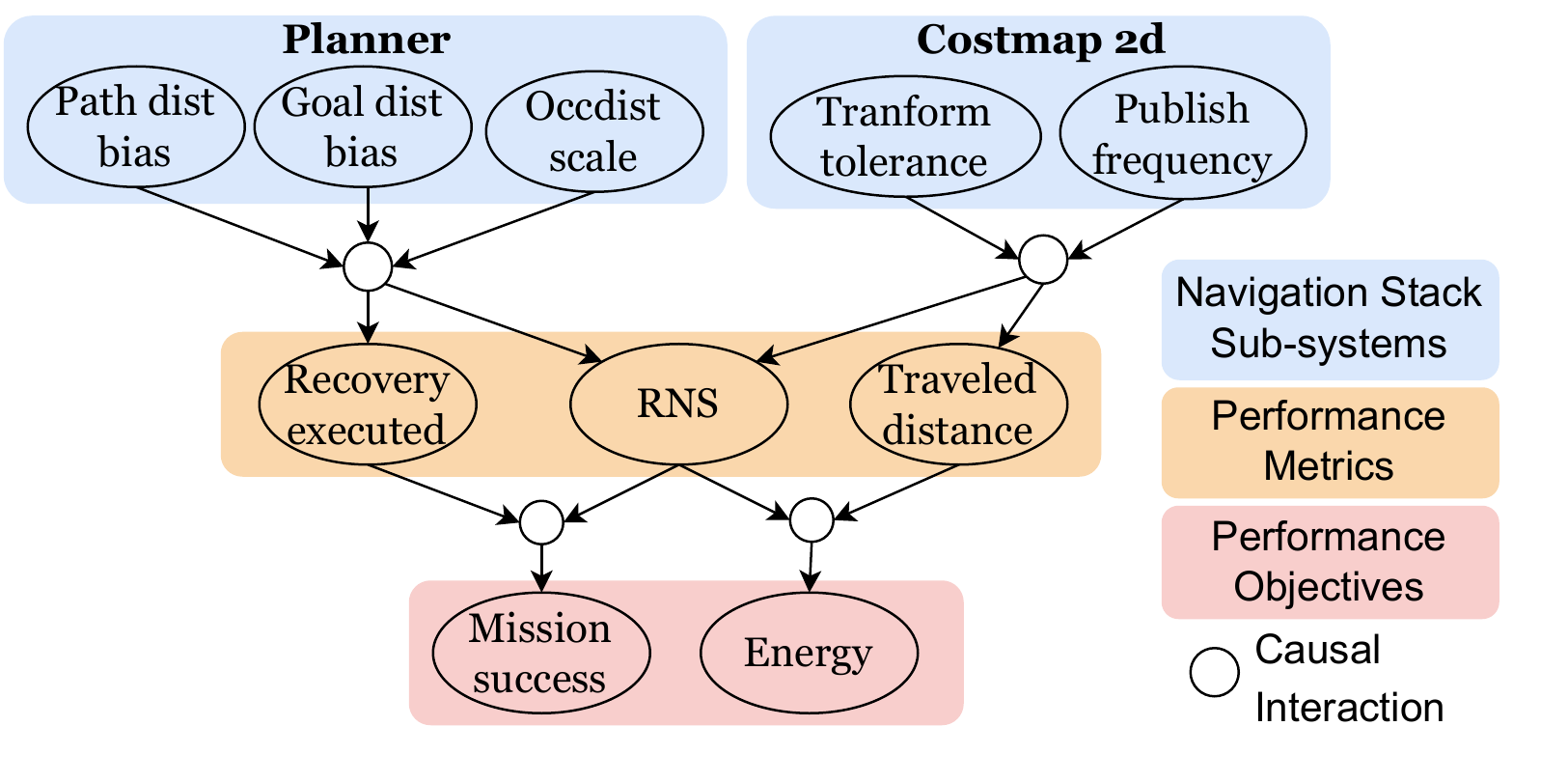}\label{fig:fci}}
   \hfill  
  \subfloat[Energy]{\includegraphics[width=.32\columnwidth]{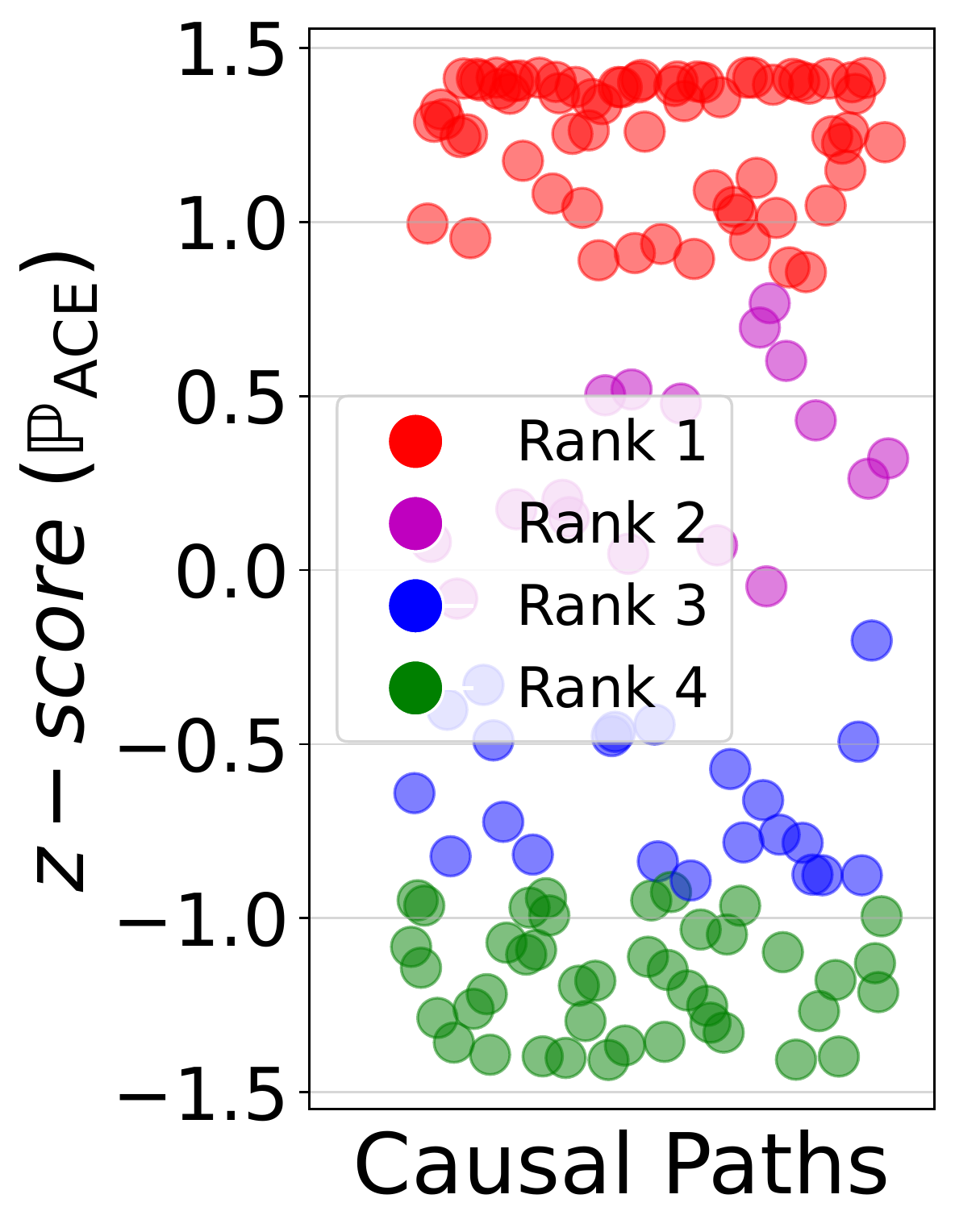}\label{fig:rank_energy}}
  \hfill
  \subfloat[Mission succes]{\includegraphics[width=.29\columnwidth,]{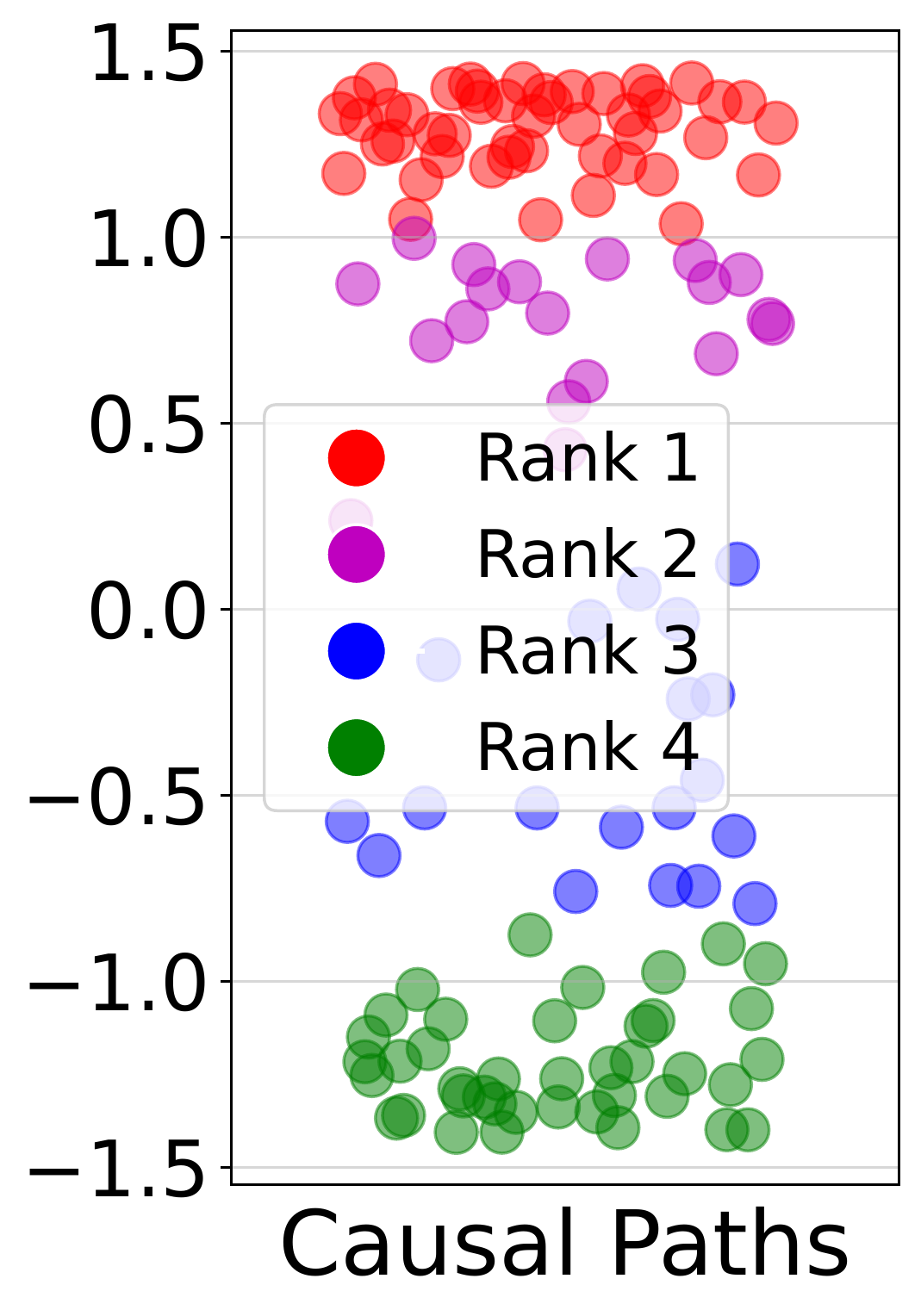}\label{fig:rank_ms}}
  \hfill
  \subfloat[Sets of options]{\includegraphics[width=0.38\columnwidth,]{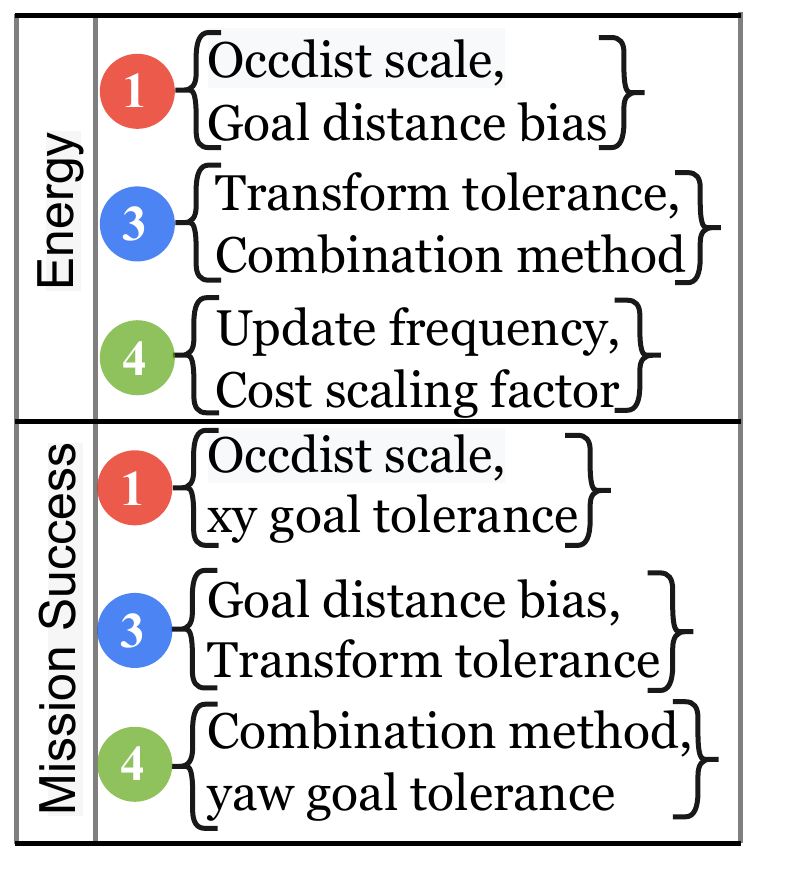}\label{fig:rank_config}}
  \caption{\small{Ranking configuration options applying Algorithm~\ref{alg:CPWE}.}\label{fig:cp_rank}}
  \vspace{0.8em}
\end{figure}

\paragraph{Learning causal model} 
We train the causal model using Algorithm~\ref{alg:cpm} on observational data obtained by running a mission 400 times under different configuration settings~(\S\ref{sec:config_options}). A partial causal model resembles the one in Fig.~\ref{fig:fci}. The arrows denote the assumed direction of causation, whereas the absence of an arrow shows the absence of direct causal influence between variables such as configuration options, performance metrics, and performance objectives. Next, we compute the $\mathbb{P}_{ACE}$ for each causal path from the causal mode~(Fig.~\ref{fig:fci}) on the performance objectives using Algorithm~\ref{alg:CPWE}. The rank of the causal paths is depicted in Fig.~\ref{fig:rank_energy} and Fig.~\ref{fig:rank_ms}. Parameters that achieve a higher rank are likely to have spurious values, hence the root cause of the functional fault. We selected two configuration options from Rank~1, Rank~3, and Rank-4 and defined three sets~(Fig.~\ref{fig:rank_config}). In our experiment, Rank~2 was discarded to demonstrate the distribution's variance because the values of $\mathbb{P}_{ACE}$ (for Rank~2) are too close to the Rank~1 and Rank~3.

\begin{figure}[!t]
\vspace{-1em}
  \centering
  \subfloat[Energy]{\includegraphics[width=0.45\columnwidth]{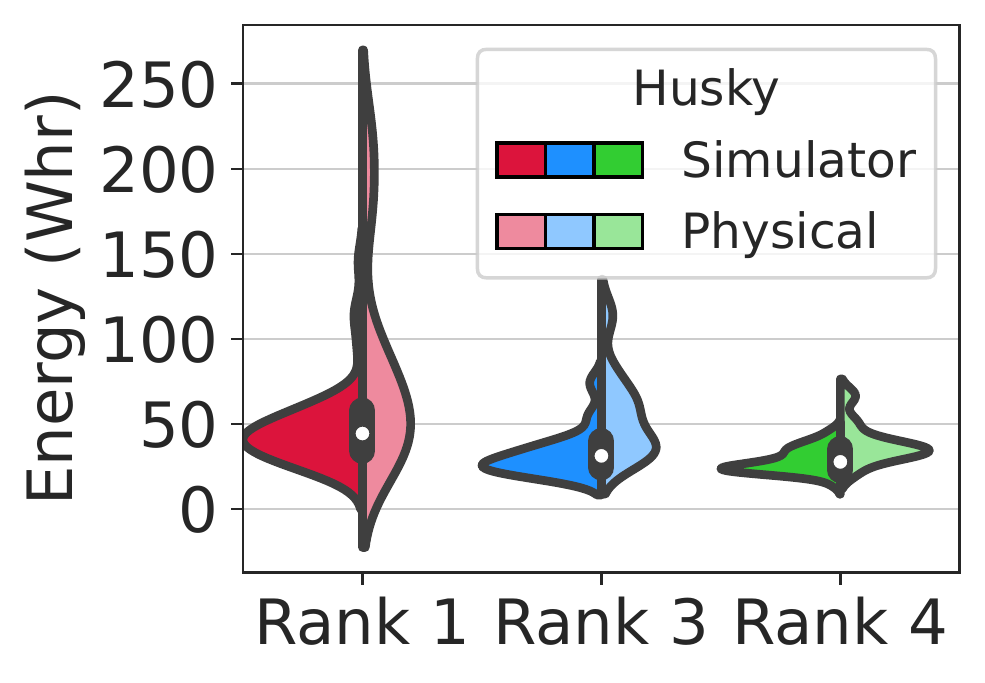}\label{fig:exp_e_sim}}
  \hfill
  \subfloat[Probability of mission success]{\includegraphics[width=0.55\columnwidth]{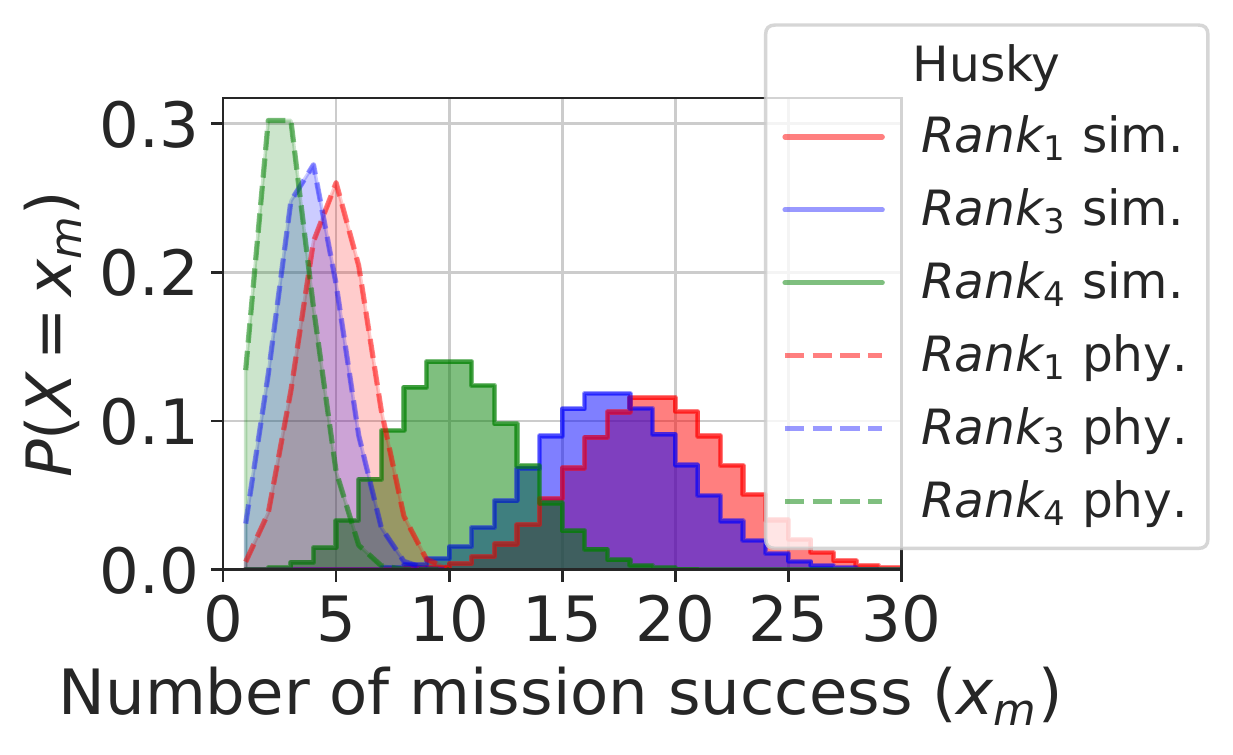}\label{fig:exp_ms_sim}} 
  \vspace{-1.2em}
  \subfloat[Traveled distance]{\includegraphics[width=.33\columnwidth,]{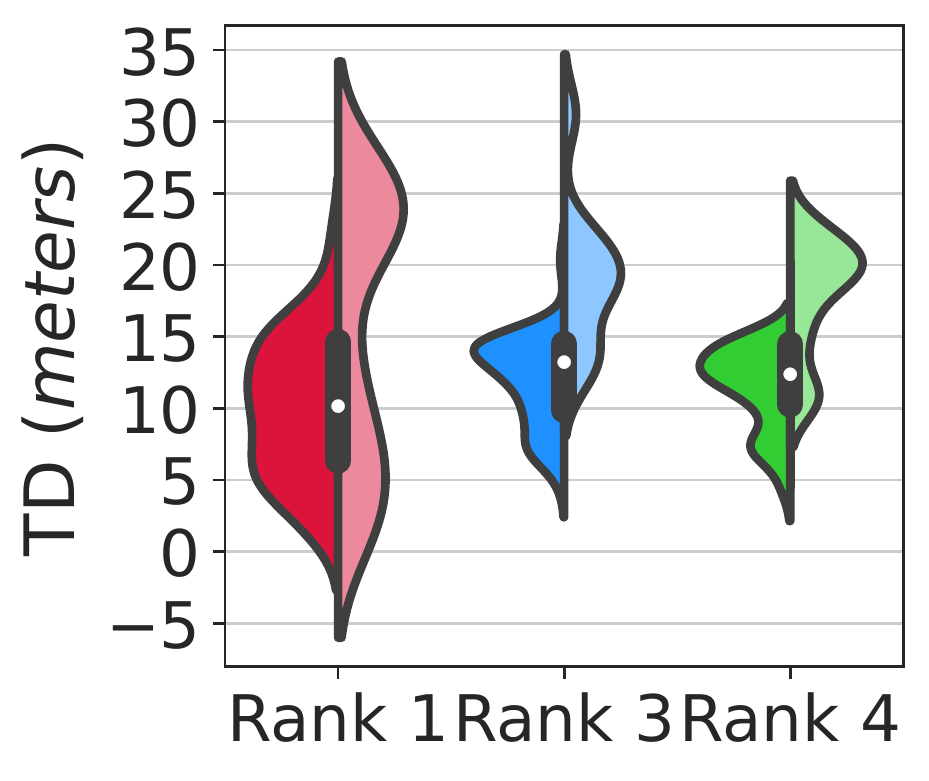}\label{fig:exp_td_sim}}
  \hfill
  \subfloat[Replanned paths]{\includegraphics[width=.33\columnwidth,]{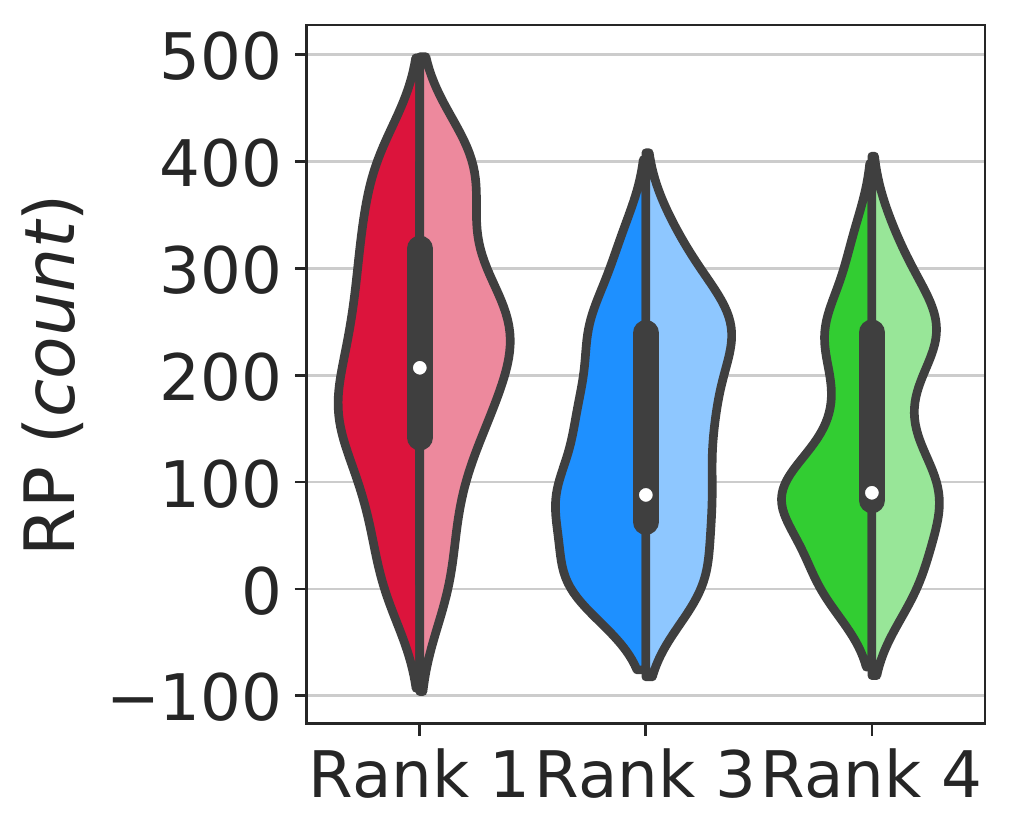}\label{fig:exp_dwa_sim}}  
  \hfill  
  \subfloat[Recovery executed]{\includegraphics[width=.33\columnwidth,]{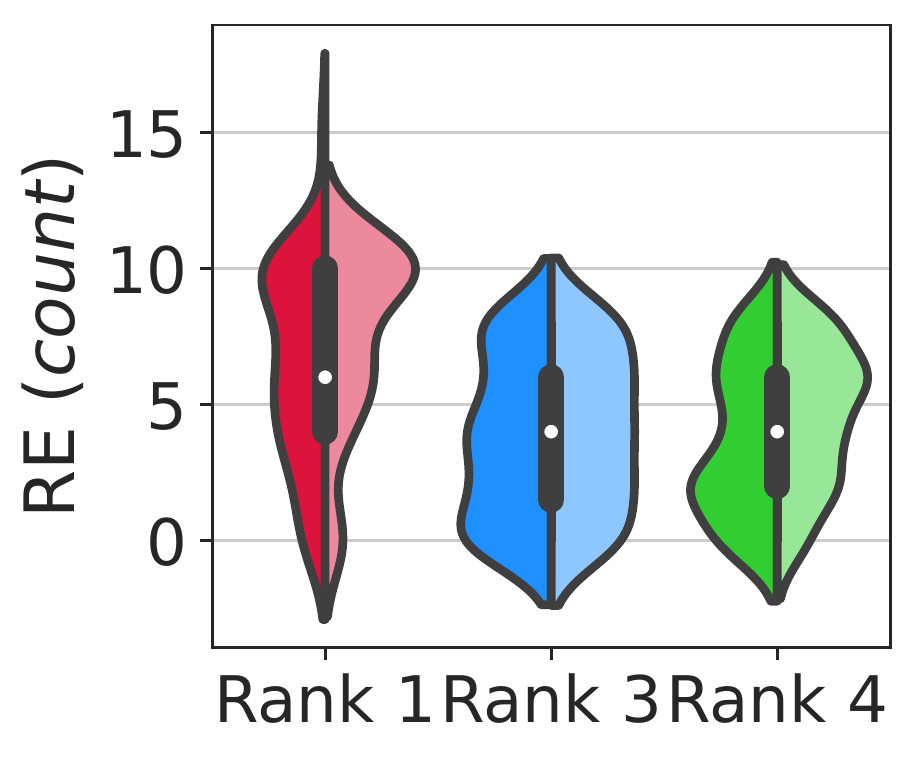}\label{fig:exp_re_sim}}  
%  \hfill
  \vspace{-1em}
  \subfloat[RNS]{\includegraphics[width=0.33\columnwidth]{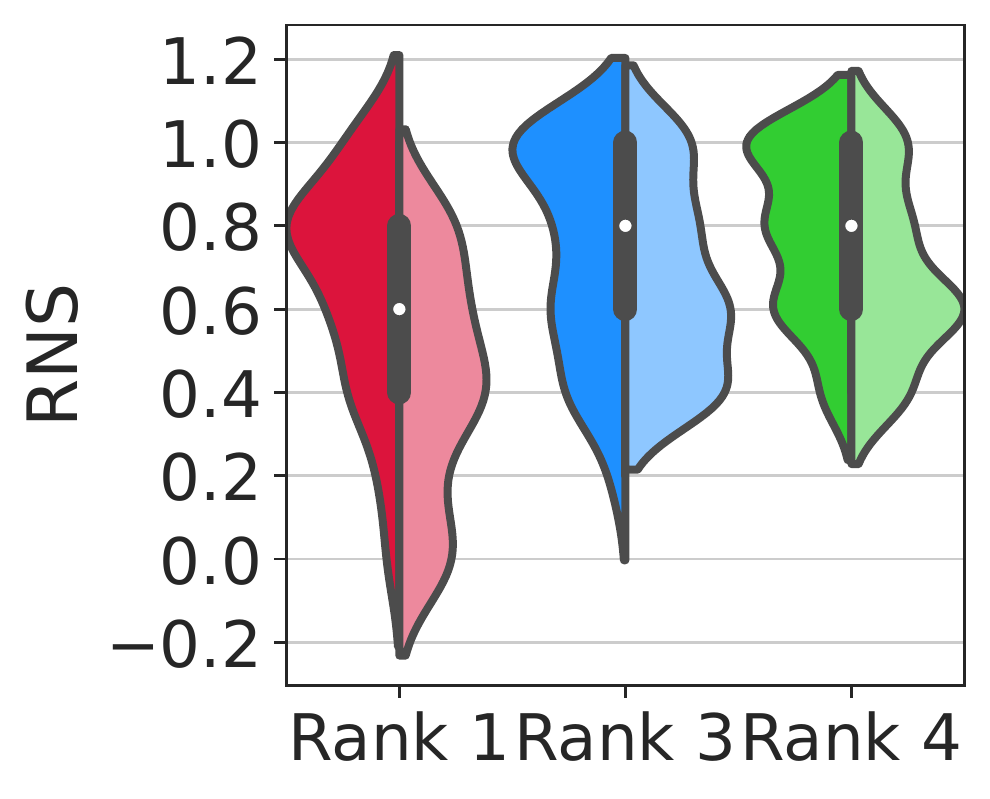}\label{fig:exp_ms_rns_sim}}  
  \hfill
  \subfloat[Recovery executed]{\includegraphics[width=.33\columnwidth,]{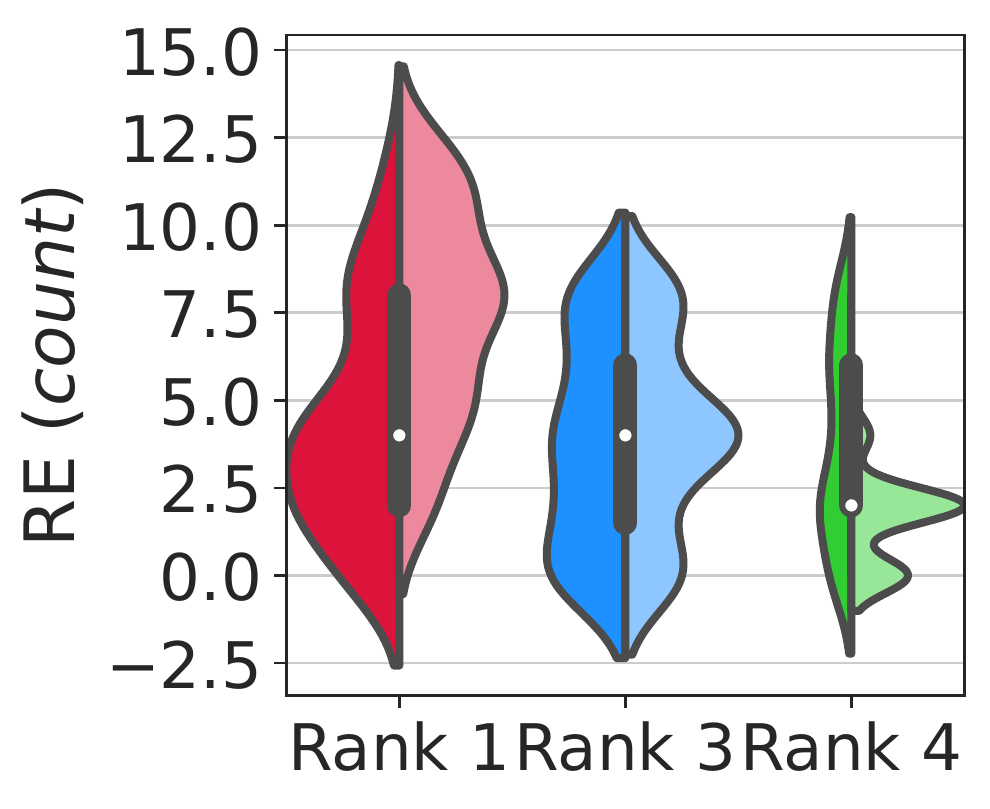}\label{fig:exp_re_ms_sim}}
  \hfill
  \subfloat[Error rotating to goal]{\includegraphics[width=.33\columnwidth,]{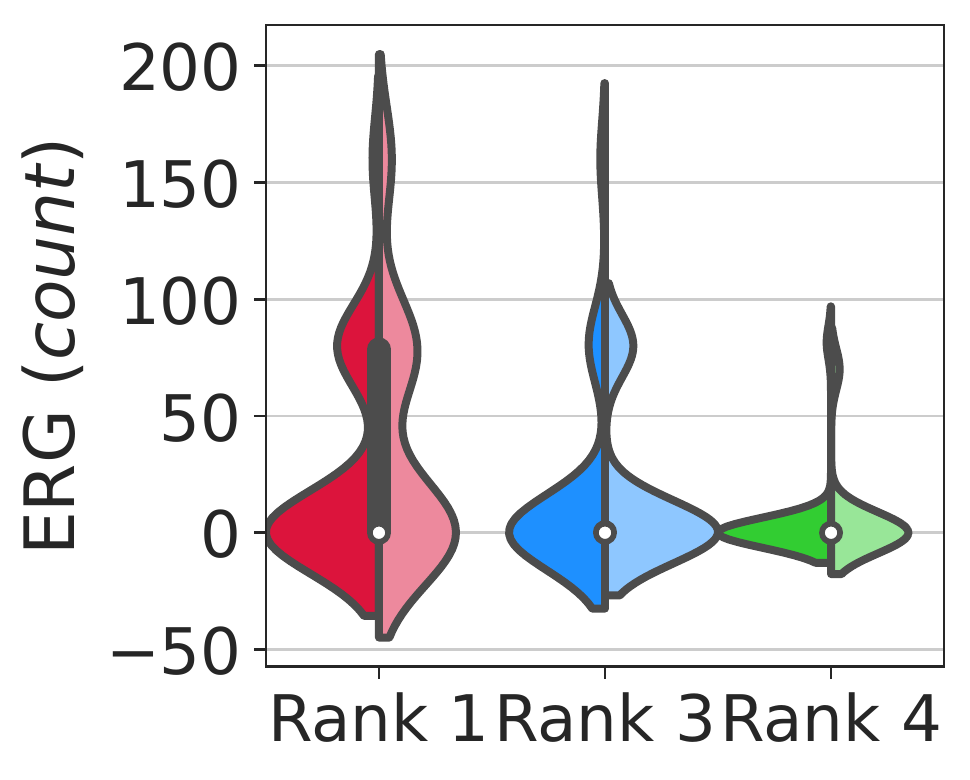}\label{fig:exp_erg_ms_sim}}
  \caption{\small{Comparing the distribution of different ranks for \label{fig:ms}
energy~(a) and mission success~(b), where (c), (d), (e) represents performance metrics of energy, and (f), (g), (h) represents performance metrics of mission success.}} \label{fig:rank_dist}
\vspace{-0.5em}
\end{figure}

\paragraph{Validating root causes diagnosed by \textsc{CaRE}}\label{sec:root cause verification}
We conducted 50 trials for each rank and recorded the energy, mission success, and performance metrics by altering only those parameter values contained in the sets (Fig.~\ref{fig:rank_config}) while leaving all other parameters to their default values (\S\ref{sec:verify_root_cause}). For instance, from Rank~1, we only changed the values of {\texTTT{occdist scale}} and {\texTTT{goal distance bias}}. Fig.~\ref{fig:rank_dist} shows violin plots to demonstrate the distribution of the trails for each rank, where the width of each curve corresponds with the frequency of $y-$\textit{axis} values and a box plot is included inside each curve. During optimization or debugging, we aimed to prioritize the configuration options which had the strongest influence on the performance objective (e.g., to optimize energy, ACE of {\texTTT{occdist scale}}~$>$~{\texTTT{transform tolerance global}}~$>$~{\texTTT{update frequency}}). Fig.~\ref{fig:exp_e_sim} and Fig.~\ref{fig:exp_ms_sim} shows that for both energy and mission success, $\sigma^2_{rank_1} > \sigma^2_{rank_3} > \sigma^2_{rank_4}$. The performance metrics are the confounding variables that influence the performance objectives (e.g., traveled distance\edgeone energy, RNS\edgeone mission success) and can be treated as the performance variance indicators. For instance, $\text{Rank~1}~:~\sigma^2_{TD}~>~\text{Rank~3}~:~\sigma^2_{TD}~>~\text{Rank-4}~:~\sigma^2_{TD}$ causes~$\uparrow\text{Rank~1}:\sigma^2_{energy}$. Similarly, for mission success (Fig.~\ref{fig:ms}), $\text{Rank~1}:\sigma^2_{RNS} > \text{Rank~3}:\sigma^2_{RNS} >\text{Rank-4}:\sigma^2_{RNS}$ causes $\downarrow \text{Rank-4}:\sigma^2_{mission\ success}$. Table~\ref{tab:compare_rank} summarizes the variance for different ranks achieved using the \textit{Husky} platform. We observe that, $\sigma^2_{rank_1}~>\sigma^2_{rank_3}~>\sigma^2_{rank_4}$ for all performance metrics and performance objective, both in the \textit{Husky} simulator and physical robot, demonstrating that configuration options which rank higher (Fig.~\ref{fig:rank_config}) have the strongest influence on the performance objectives. Moreover, \textsc{CaRE} achieved $95\%$ accuracy when comparing the predicted root causes with the ground truth data~(Fig.\ref{fig:care_acc}).

\subsection{RQ2: Transferability} \label{sec:transferability}
The configuration options that specify the hardware characteristics of the physical platform differ across robotic systems~(e.g., sensor frequency), and these hardware characteristics can significantly impact the performance of the tasks carried out by the robotic systems. We answer RQ2 by reusing the causal model in a different robotic platform. In particular, we reuse the causal model constructed from a source platform, e.g., the \textit{Husky} simulator (Fig.~\ref{fig:fci}), to determine the root causes of a functional fault in a target platform, e.g., \textit{Turtlebot~3}. 
\begin{table}[!t]
\centering
\caption{Comparison of the variance ($\sigma^2$) of different ranks for energy and mission success using the \textit{Husky} platform.}
\scalebox{0.95}{
\begin{tabular}{c|c|l|lll|} 
\cline{4-6}
\multicolumn{1}{l}{} & \multicolumn{1}{l}{} & \multicolumn{1}{c|}{} & \multicolumn{1}{c}{$\sigma^2_{rank_1}$} & \multicolumn{1}{c}{$\sigma^2_{rank_3}$} & \multicolumn{1}{c|}{$\sigma^2_{rank_4}$} \\ 
\hhline{---===}
\multirow{8}{*}{\rotatebox{90}{Husky simulator}} & \multicolumn{1}{l|}{\multirow{2}{*}{Objective}} & Energy & 24.32 & 13.78 & 7.27 \\
 & \multicolumn{1}{l|}{} & Mission success & 11.77 & 11.22 & 8.01 \\ 
\cline{2-6}
 & \multirow{4}{*}{\rotatebox{90}{Energy}} & Traveled distance & 4.93 & 3.14 & 2.90 \\
 &  & Replanning path & 123.39 & 100.97 & 97.73 \\
 &  & Recovery executed & 3.63 & 2.88 & 2.70 \\ 
 &  & Mission time & 46.96 & 29.21 & 20.93 \\  
\cline{2-6}
 & \multirow{4}{*}{\rotatebox{90}{\begin{tabular}[c]{@{}c@{}}Mission\\success\end{tabular}}} & RNS & 0.26 & 0.25 & 0.20 \\
 &  & Recovery executed & 3.21 & 2.94 & 2.79 \\
 &  & Error rotating to goal & 44.01 & 40.59 & 16.13 \\
 &  & Mission time & 153.93 & 57.75 & 38.85 \\ 
\hline\hline
\multicolumn{1}{l|}{\multirow{8}{*}{\rotatebox{90}{Husky physical}}} & \multirow{2}{*}{Objective} & Energy & 64.43 & 26.57 & 12.43 \\
\multicolumn{1}{l|}{} &  & Mission success & 2.46 & 2.17 & 1.60 \\ 
\cline{2-6}
\multicolumn{1}{l|}{} & \multirow{4}{*}{\rotatebox{90}{Energy}} & Traveled distance & 9.84 & 5.22 & 4.23 \\
\multicolumn{1}{l|}{} &  & Replanning path & 126.38 & 109.07 & 108.87 \\
\multicolumn{1}{l|}{} &  & Recovery executed & 3.50 & 2.98 & 2.67 \\ 
\multicolumn{1}{l|}{} &  & Mission time & 34.39 & 25.90 & 16.93 \\ 
\cline{2-6}
\multicolumn{1}{l|}{} & \multirow{4}{*}{\rotatebox{90}{\begin{tabular}[c]{@{}c@{}}Mission\\success\end{tabular}}} & RNS & 0.29 & 0.23 & 0.21 \\
\multicolumn{1}{l|}{} &  & Recovery executed & 3.17 & 2.83 & 1.26 \\
\multicolumn{1}{l|}{} &  & Error rotating to goal & 56.11 & 33.59 & 22.14 \\
\multicolumn{1}{l|}{} &  & Mission time & 62.85 & 42.43 & 35.39 \\
\hline
\end{tabular}
}
\label{tab:compare_rank}
\vspace{1em}
\end{table}

\paragraph{Ground truth} We measured 400 samples (performance metrics), varying the configuration options for both \textit{Husky} in simulation and \textit{\textit{Turtlebot~3}}. We curated a ground truth of functional faults using the ground truth data. In particular, we curated the ground truth for the two performance objectives: (i) configurations that result in a mission failure, and (ii) configurations that achieved energy consumption worse than the $99^{th}$ percentile are labeled as `\textit{faulty}' (non-functional fault). The ground truth contains 20 functional faults (10~mission success, 10~energy), and each has two to four root causes.

\paragraph{Baseline} We compared \textsc{CaRE} against the state-of-the-art Cooperative Bug Isolation~(CBI)~\cite{liblit2005scalable}---a statistical debugging method that uses a feature selection algorithm. We selected CBI for its use of statistical methods similar to ours and its ability to identify multiple root causes. However, unlike our approach, CBI relies on correlations instead of causation to identify the root causes of the fault. We computed the \textit{Importance} \textit{score}~\cite{liblit2005scalable} by computing $Failure(P)$, $Context(P)$, and $Increase(P)$ for different configuration options and objectives. Based on the importance score, we ranked the configuration options similarly to Fig.~\ref{fig:cp_rank}. In our experiments, we set the confidence intervals to $95\%$ to eliminate configuration options with low confidence due to few observations but a high $Increase(P)$.

\begin{figure}[!t]
\vspace{-0.7em}
  \centering
  \subfloat[Experimental environment]{\includegraphics[width=.57\columnwidth]{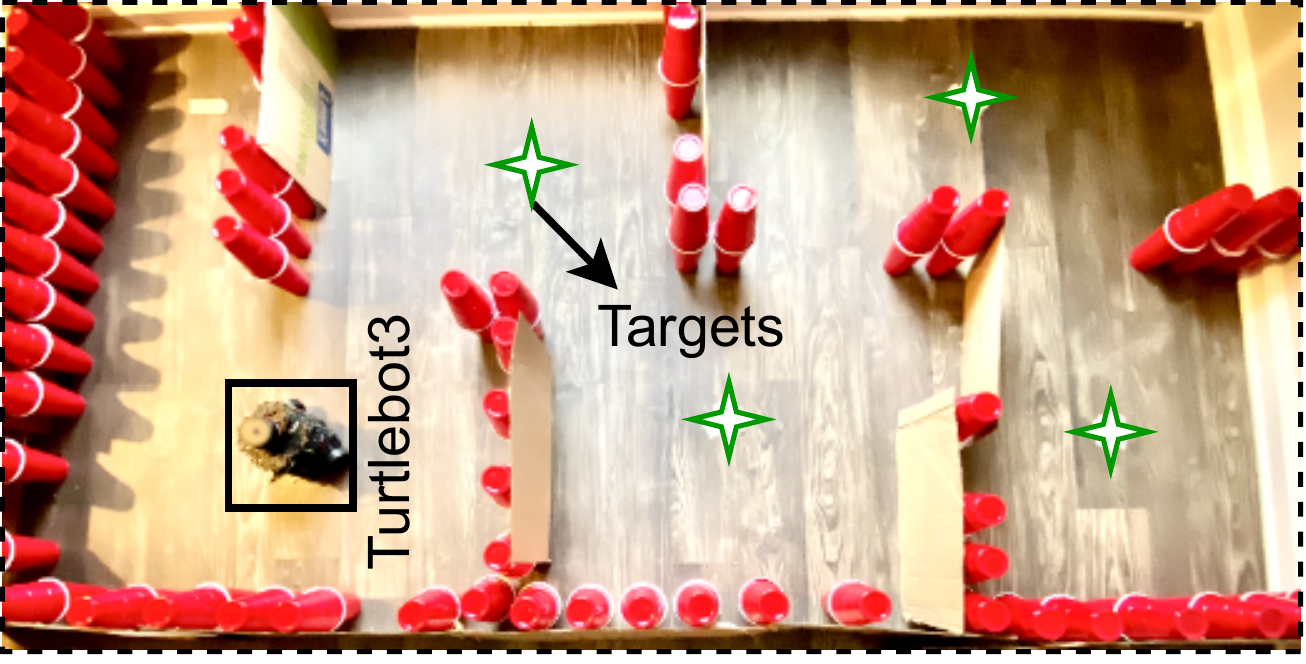}\label{fig:turtlebot_env}}
  \hfill
  \subfloat[Generated map using SLAM: {\texTTT{gmapping}}]{\vstretch{1.15}{\includegraphics[width=0.42\columnwidth]{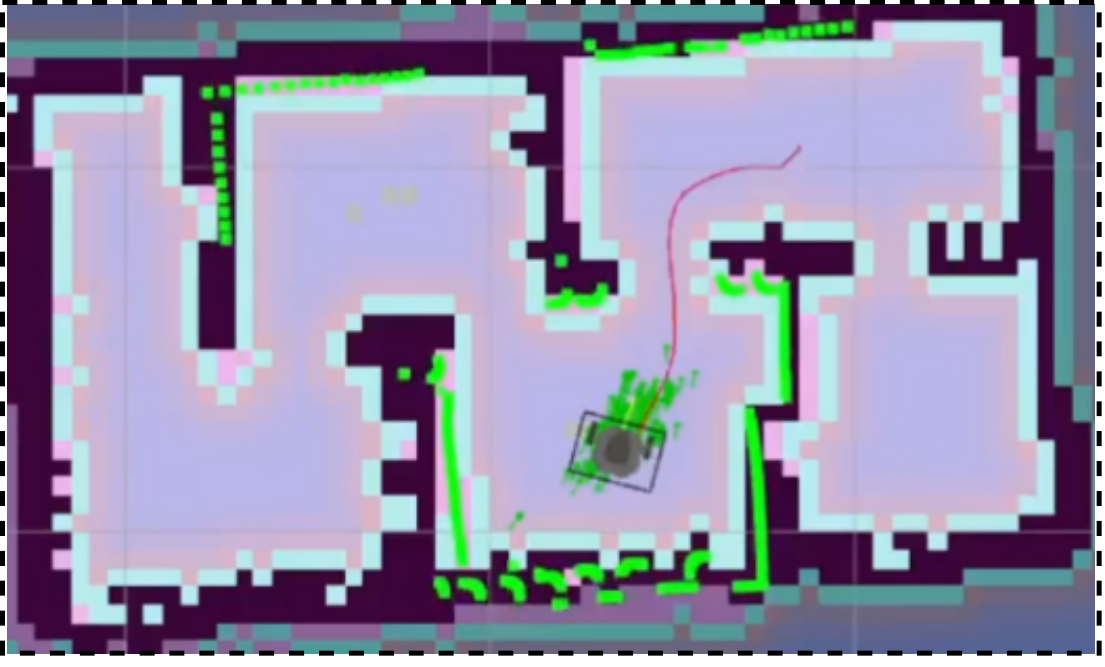}\label{fig:turtlebot_map}}}
  \caption{\small{Experimental environment for \textit{Turtlebot~3} composed of five target locations and narrow passageways.}}
  \label{fig:turtlebot_exp_env}
   \vspace{-1em} 
\end{figure}

\paragraph{Setting} We use \textit{Turtlebot~3}~\cite{turtlebot3} as our target platform. Using Gmapping SLAM~\cite{turtlebot3} node, we generate the map~(Fig.~\ref{fig:turtlebot_map}) of the experimental environment~(Fig.~\ref{fig:turtlebot_env}). We direct the robot to autonomously navigate to five target locations using {\texTTT{turtlebot3\_navigation.launch}}.  If the \textit{Turtlebot} robot reached each of the five target locations, the mission was considered successful. We measure and collect the performance data while the \textit{Turtlebot} performs a mission using the interface shown in Fig.~\ref{fig:data_collect}.

\paragraph{Results} Given a set of test data, ground truth, and \textsc{CaRE}'s predictions on the test data, we evaluated the predictions by dividing them into true and false positives and negatives (\textit{TP, FP, TN,} and \textit{FN}). Subsequent metrics include:
\begin{itemize}
    \item \textit{Accuracy}: The measure of the predicted root causes that match the ground truth root causes,~$(TP+TN)/(TP+FP+TN+FN)$.
    \item \textit{Precision}: The ratio of true root causes among the predicted ones, $TP/(TP+FP)$.
    \item \textit{Recall}: The ratio of true root causes that
    are correctly predicted, $TP/(TP+FN)$.
    \item \textit{F1-score}: The harmonic mean of precision and recall, $2 \times (\mathrm{precision} \times \mathrm{recall}) / (\mathrm{precision} + \mathrm{recall})$.
    \item \textit{RMSE}: Weighted difference between the predicted and true root causes. For example, if $\hat{y}$ is the predicted root cause of a functional fault and $y$ is root cause in the ground truth, we measure \textit{RMSE}~$=\sqrt{\frac{1}{N}\sum_{i=1}^{N}({\mathrm{ACE}(y)} - {\mathrm{ACE}(\hat{y})})^2}$, where ACE is computed using Equation~\ref{eq:ace}.
\end{itemize}

\begin{figure}[!t]
\vspace{-0.5em}
  \centering
  \subfloat[Baseline]{\includegraphics[width=0.23\columnwidth]{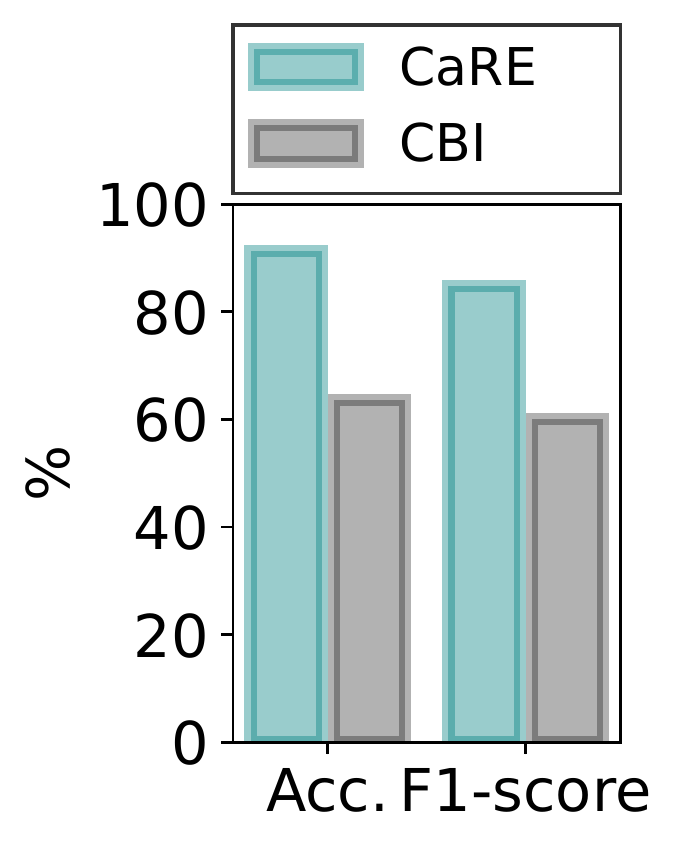}\label{fig:baseline_compare}}  
   \hfill
  \subfloat[\textsc{CaRE} accuracy]{\includegraphics[width=.44\columnwidth]{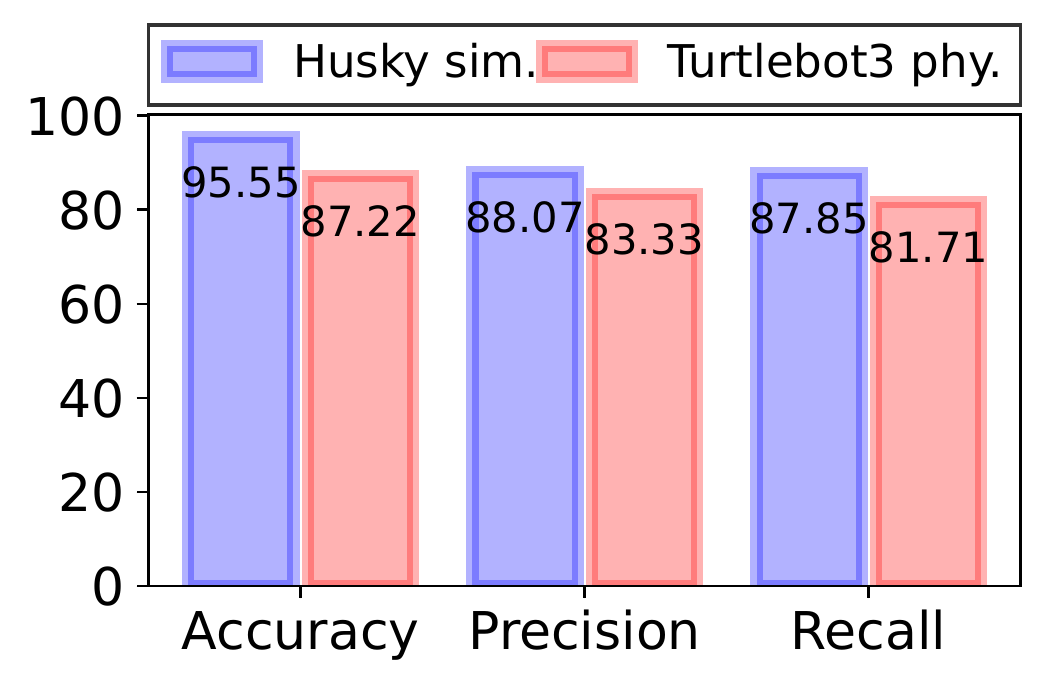}\label{fig:care_acc}}
  \hfill
  \subfloat[\textsc{CaRE} error]{\includegraphics[width=0.33\columnwidth]{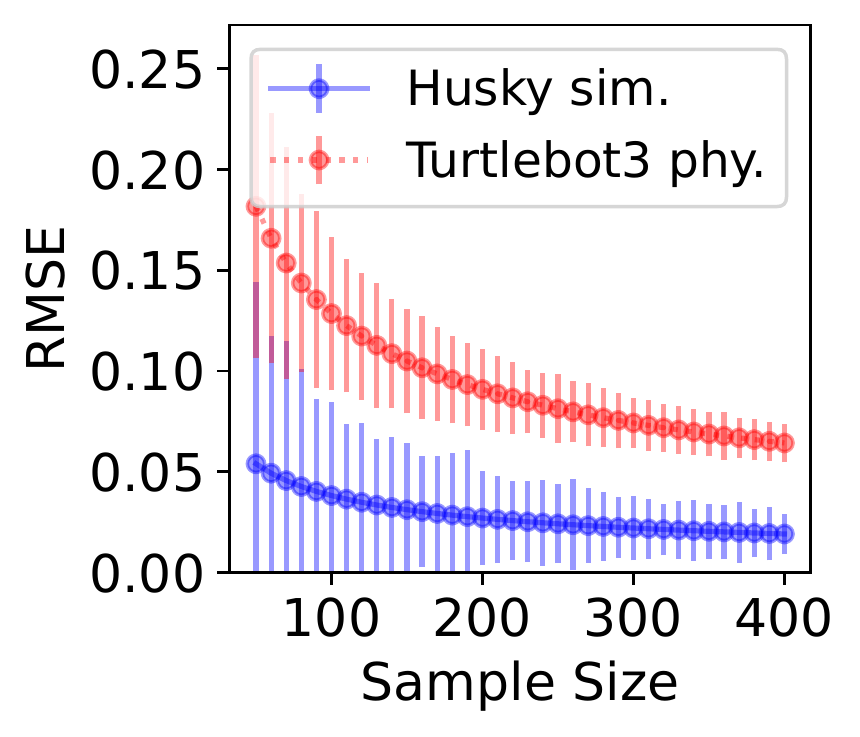}\label{fig:care_rmse}}
  \caption{\small{Comparing \textsc{CaRE} against CBI (a),} and demonstrating \textsc{CaRE}'s transferability (b), (c) by reusing the causal model constructed from \textit{Husky} in simulation, to diagnose the root causes of the functional faults in the \textit{Turtlebot~3} physical robot.}
  \label{fig:care_transferability}
  \vspace{0.5em}
\end{figure}

\noindent Fig.~\ref{fig:care_transferability} shows the results in diagnosing the root causes of the mission success and energy faults. The total \textit{accuracy} is computed using $\frac{1}{N_{obj}}\sum_{i=1}^{N_{obj}}acc.$, where $N_{obj}$ is the number of performance objectives; similarly for \textit{precision} and \textit{recall}. \textsc{CaRE} achives $27\%$ more accuracy, and $24\%$ more F1-score compared to CBI~(Fig.~\ref{fig:baseline_compare}). We computed the \textit{accuracy} in Fig.~\ref{fig:baseline_compare} using $(\mathrm{Husky}_{total\_acc.} + \mathrm{Turtlebot3}_{total\_acc.})/2$; similarly for F1-score. We observe that reusing \textsc{CaRE} in \textit{Turtlebot~3} obtains $8\%$ less accuracy, $4\%$ less precision, and $6\%$ less recall compared to the source platform (\textit{Husky} simulator). We also observe higher \textit{RMSE} in the \textit{Turtlebot~3} platform (the total \textit{RMSE} is computed using $\sum_{i=1}^{N_{obj}}RMSE$). However, if we increase the sample size, \textsc{CaRE} incrementally updates the internal causal model with new samples from the target platform  to learn the new relationships, and we observe a decrease in \textit{RMSE}~(see Fig.~\ref{fig:care_rmse}). Therefore, the model does transfer reasonably well.
\section{Discussion}
\paragraph{Usability of \textsc{CaRE}}
The design we have proposed is general and extendable to other robotics systems, but would require some engineering efforts. In particular, to add a new variable to the causal model, the following steps would be required: (i)~identifying the manipulable and non-manipulable variables, (ii)~profiling the observational data related to the new variable, including its corresponding performance objectives, (iii)~learning and adding the causal relationships of the new variable to the existing model. Furthermore, to support a new robotic system, in addition to step~1, profiling the observational data for the entire configuration space would be required to train the causal model. We provide a tool for this in \S\ref{sec:reval}, but it is currently limited to ROS-based systems. Although the effort is non-trivial, it is required one time for each robotic system. For instance, it took about 8$\sim$12 hours applying \textsc{CaRE} to Turtlebot~3. 

\paragraph{Why did \textsc{CaRE} outperform CBI?}
\textsc{CaRE} discovers the root causes of the configuration bugs by learning a causal model that focuses on the configurations that have the highest causal effect on the performance objectives, eliminating the irrelevant configuration options. For instance, while finding the root causes of the functional and non-functional faults on the performance objectives, CBI reported $116$ FP, whereas \textsc{CaRE} reported only $13$ FP (\textit{Huksy} and \textit{Turtlebot}~\textit{3} combined), hence, achieving a higher F1-score compared to CBI (Fig.~\ref{fig:baseline_compare}). CBI reported a higher number of FP because it determines the root causes based on the correlation between variables. For instance, it identified planner failure rates increase the $P${\texTTT{(mission success)}}, which is counter-intuitive. Therefore, an engineer would spend less time debugging and optimizing the parameters when using \textsc{CaRE}.

\paragraph{Dealing with partially resolved causal graphs}
When there is insufficient data and edges are returned that are partially directed, existing off-the-shelf causal graph discovery algorithms like FCI remain ambiguous. Obtaining high-quality data for highly configurable robotic systems is challenging. To address this issue, we combine FCI with an information-theoretic approach using entropy (we provide Algorithm~\ref{alg:cpm} for this). This strategy reduces ambiguity, allowing the causal graph to converge faster. Note that establishing a theoretical guarantee for convergence is out of scope of this paper because having a global view of the entire configuration space is impractical. Moreover, the causal model's correctness can be affected if there are too many unmeasured confounders, and this error may propagate along with the structure if the dimensionality is high.

\paragraph{Limitations} 
The efficacy of \textsc{CaRE} depends on several factors, including the representativeness of the observational data and the presence of unmeasured confounders, which deteriorate the accuracy. In some cases, the causal model may be missing some important connections, resulting in identifying spurious root causes. One promising direction to address this problem would be to develop novel approaches for learning better structure by incorporating domain knowledge to define constraints on the structure of the underlying causal model. In addition, there is potential for developing better sampling algorithms by searching the space more efficiently to discover effective interventions that allow for faster convergence to the true underlying structure.

\paragraph{Future directions}
We envision two possible avenues for our paper: empirical and technical. For the empirical aspect, \textsc{CaRE} could be applied to improve autonomy in robotic spacecraft missions and more complicated scenarios (e.g., coordinated multi-robot exploration---a team of robots distributed in an environment, to solve complex tasks) can be the future work. The technical aspect could involve performing static analyses to extract the configurable parameters from a robotic system in an automated manner.
\section{Related work}
\paragraph{Debugging using performance influence model}
Prior work on highly configurable systems has revealed that the majority of performance issues are related to configuration space~\cite{HanEmpirical2016}. Previous approaches have used performance influence models~\cite{VelezConfigCrusher2020,ChenPerformance2022,VelezDebugging2022,CaoUnderstanding2022,9401975,muhlbauer2020identifying,HaochenTuning2019,8952290} to discover and locate performance bugs in software systems. The majority of these approaches used machine learning setting such as Gaussian processes~\cite{8952290}, linear (LASSO) regression~\cite{muhlbauer2020identifying}, classification and regression trees (CART)~\cite{9401975} to model configuration options as features and learn a corresponding prediction function. 

\paragraph{Debugging misconfigurations in robotic systems}
Researchers often use random testing such as fuzzing~~\cite{KimRoboFuzz2022,9561627,SeulbaeDriveFuzz2022,9811701} and delta debugging~\cite{von2021automated,7353866,8449261} approaches to debug and enhance robotic systems' performance. Moreover, several studies have proposed different methods to deal with the configuration bugs, such as misconfiguration debugging~\cite{jung2021swarmbug}, automatic parameter tuning~\cite{8981601,9561311}, discovering and fixing configuration bugs in co-robotic systems~\cite{taylor2016co}, and  statically identifying run-time architectural misconfigurations~\cite{9779703}. Data-driven machine learning techniques have also been widely applied to improve performance by fine-tuning configuration parameters or diagnosing misconfigurations, as opposed to heavily relying on human expertise. For instance, machine learning techniques, including Gaussian processes~\cite{9197260}, artificial neuro-Fuzzy inference~\cite{electronics8090935}, and reinforcement learning~\cite{XIAO2022104132}, are used for parameter tuning in various subsystems of the ROS navigation stack. However, these techniques may not be effective at applying knowledge in different environments and may have difficulty retaining past information~\cite{9345478}.

\paragraph{Causal learning for systems} Machine learning techniques have been proven effective at identifying correlations in data, though they are ineffective at identifying causes~\cite{pearl2009causality}. To address this challenge, several studies, including detecting and understanding the defect's root causes~\cite{johnson2020causal, DubslaffCausality2022}, improving fault localization~\cite{9402143}, and reasoning about system's performance~\cite{iqbal2022unicorn}, utilize causal learning. Using the encoded information, we can benefit from analyses that are only possible when we explicitly employ causal models, in particular, interventional and counterfactual analyses~\cite{pearl2009causality, spirtes2000causation}. More recently, Swarmbug~\cite{jung2021swarmbug}, a method for debugging configuration bugs in swarm robotics, utilized a causality-based approach to find and fix the misconfigurations in swarm algorithms. However, the Swarmbug method is specifically designed for use in swarm robotics and is therefore only useful for diagnosing configuration bugs in swarm algorithms.
\section{Conclusion}
We proposed \textsc{CaRE}, a novel approach to determining the root causes of functional faults in robotic systems. \textsc{CaRE} learns and exploits the robotic system’s causal structure consisting of manipulable variables~(configuration options), non-manipulable variables~(performance metrics), and performance objectives. Then, given the causal model, \textsc{CaRE} extracts the paths that lead from configuration options to the performance objectives and determines the configuration options that have the highest causal effect on the performance objective by computing the average causal effect of each path. Our evaluation shows that \textsc{CaRE} effectively diagnoses the root cause of functional faults, and the learned causal model is transferable across different robotic systems.
\section*{ACKNOWLEDGMENT}
Part of this research was carried out at the Jet Propulsion Laboratory, California Institute of Technology, under a contract with the National Aeronautics and Space Administration (80NM0018D0004). In addition, this work has partly been supported by  National Science Foundation~(Awards 2233873, 2007202, and 2107463) and the Spanish Government (FEDER/MICINN-AEI) under projects TED2021-130523B-I00 and PID2021-125527NB-I00. We thank Hari Nayar, Michael Dalal, Ashish Goel, Erica Tevere, Anna Boettcher, Anjan Chakrabarty, Ussama Naal, Carolyn Mercer, Issa Nesnas, Matt DeMinico, Md Shahriar Iqbal, and Jianhai Su for contributions to the \textsc{CaRE} framework and evaluations on the NASA testbeds: \url{https://nasa-raspberry-si.github.io/raspberry-si/}. 

\bibliographystyle{IEEEtran}
\bibliography{main_arXiv}

\clearpage
\section*{Artifact Appendix} \label{sec:sup-mat}
\begin{tcolorbox}[colback=blue!5!white,colframe=blue!75!black]
\textbf{DOI:} \url{doi:10.5281/zenodo.7529716} \\
\textbf{Code:} \url{https://github.com/softsys4ai/care}
\end{tcolorbox}
This appendix provides additional information regarding \textsc{CaRE}. We describe the steps to reproduce the results reported in \S\ref{sec:exp} using \textsc{CaRE}. The source code and data are provided in a publicly accessible GitHub repository, allowing users to test them on any hardware once the software dependencies are met.

\subsection{Profiling the Observational Data} \label{sec:reval}
To collect the observational data, we developed a profiling tool, \href{https://semopy.com/}{\color{blue!80}\textit{Reval}}, which currently supports \textit{Husky} and \textit{Turtlebot-3}. Note that, the following steps are optional since we provide the previously measured observational Data. Additionally, Ubuntu-20.04, and \href{http://wiki.ros.org/noetic/Installation/Ubuntu}{\color{blue!80}Ros Noetic} are prerequisites for using \textit{Reval}.

\subsubsection{Software Dependencies} Install the software dependencies for \textit{Reval} using the following commands: 
\begin{lstlisting}[basicstyle=\ttfamily\scriptsize, frame=single, backgroundcolor=\color{blue!5}]
$ apt install ripgrep
$ pip install pandas
$ pip install tqdm
$ pip install tabulate 
\end{lstlisting}
\noindent Also, the dependencies can be automatically installed using the following commands:
\begin{lstlisting}[basicstyle=\ttfamily\scriptsize, frame=single, backgroundcolor=\color{blue!5}]
$ git clone https://github.com/softsys4ai/Reval.git
$ sh Reval/requirements.sh
\end{lstlisting}

\noindent Download and install {\texTTT{ros\_readbag.py}} using the following commands:
\begin{lstlisting}[basicstyle=\ttfamily\scriptsize, frame=single, backgroundcolor=\color{blue!5}]
$ wget https://raw.githubusercontent.com/ 
ElectricRCAircraftGuy/eRCaGuy_dotfiles/master/ 
useful_scripts/ros_readbagfile.py
$ sudo chmod +x ros_readbagfile.py
$ mkdir -p ~/bin
$ mv ros_readbagfile.py ~/bin/ros_readbagfile
$ ln -si "${PWD}/ros_readbagfile.py" ~/bin/ros_readbagfile
\end{lstlisting}
\noindent Note that, if this is the first time ever creating the{\texTTT{~/bin}} directory, then log out and log back in to your Ubuntu user account to cause Ubuntu to automatically add your{\texTTT{~/bin}} directory to your executable {\texTTT{PATH}}.

\subsubsection{Dependencies for \textit{Husky} simulator} Install the dependencies using the following commands:
% \begin{tcolorbox}[halign=center,colback=blue!5!white,colframe=black!50!black,boxrule=0.5pt, width=3.6in]
%\vspace{-1em}
\begin{lstlisting}[basicstyle=\ttfamily\scriptsize, frame=single, backgroundcolor=\color{blue!5}]
$ apt-get install ros-noetic-husky-simulator
$ apt-get install ros-noetic-husky-navigation
$ apt-get install ros-noetic-husky-desktop
\end{lstlisting}
% \vspace{-1em}
% \end{tcolorbox}

\subsubsection{Dependencies for \textit{Turtlebot-3} physical} Install the dependencies using the following commands:
\begin{lstlisting}[basicstyle=\ttfamily\scriptsize, frame=single, backgroundcolor=\color{blue!5}]
$ mkdir -p ~/Reval/src/turtlebot3
$ git clone -b noetic-devel https://github.com/ 
ROBOTIS-GIT/DynamixelSDK.git
$ git clone -b noetic-devel https://github.com/ 
ROBOTIS-GIT/turtlebot3_msgs.git
$ git clone -b noetic-devel https://github.com/ 
ROBOTIS-GIT/turtlebot3.git
\end{lstlisting}

\subsubsection{Installation and Running} For building \textit{Reval} from source use the following commands:
\begin{lstlisting}[basicstyle=\ttfamily\scriptsize, frame=single, backgroundcolor=\color{blue!5}]
$ git clone - b husky https://github.com/softsys4ai/Reval.git
$ source /opt/ros/noetic/setup.bash
$ cd ~/Reval && catkin build
\end{lstlisting}
\noindent For \textit{Turtlebot-3} use {\texTTT{-b turtlebot3}}.

\noindent For execution, use the following commands,
\begin{lstlisting}[basicstyle=\ttfamily\scriptsize, frame=single, backgroundcolor=\color{blue!5}]
$ source devel/setup.bash
$ python reval.py
\end{lstlisting}

\noindent Additionally, the following arguments can be used during execution:
\begin{lstlisting}[basicstyle=\ttfamily\scriptsize, frame=single, backgroundcolor=\color{blue!5}]
optional arguments:
  -h, --help    show this help message and exit
  -v , -viz     turn on/off visualization of gazebo and rviz
                (default: On)
  -e , -epoch   number of data-points to be recorded
                (default: 1)
\end{lstlisting}
\noindent For example, {\texTTT{``python reval.py -v off -e 10''}}

\subsection{Running \textsc{CaRE}}
\textsc{CaRE} is implemented by integrating and building on top of several existing tools:
\begin{itemize}
    \item \href{https://github.com/cmu-phil/causal-learn}{\color{blue!80}causal-learn} for structure learning. 
    \item \href{https://ananke.readthedocs.io/en/latest/}{\color{blue!80}ananke} and \href{https://github.com/akelleh/causality}{\color{blue!80}causality} for estimating the causal effects.
\end{itemize}

\subsubsection{Installation} To build \textsc{CaRE} from source, use the following commands:
\begin{lstlisting}[basicstyle=\ttfamily\scriptsize, frame=single, backgroundcolor=\color{blue!5}]
$ git clone https://github.com/softsys4ai/care.git
$ cd ~/care && pip install -r requirements.txt
\end{lstlisting}

\subsubsection{Data} All the datasets required to run experiments are already
included in the {\texTTT{./care/data}} directory. Additionally, the {\texTTT{./care/observationa\_data}} directory contains the required observational data to train the causal model.

\subsection{Experiments}
We run the following experiments to support our claims.

\subsubsection{E1: Root cause Verification Experiment} We first train the causal model using the observational data, and compute the ranks of the causal paths (the path's ranks are provided in the {\texTTT{./care/result/rank\_path.csv}} file). We conducted 50 trials for each rank and recorded the energy, mission success, and performance metrics both in \textit{Husky} simulator and physical robot. The result of the trails are provided in the {\texTTT{./care/result/exp}} directory. To train the causal model and compute the causal path's rank, execute the following command:
\begin{lstlisting}[basicstyle=\ttfamily\scriptsize, frame=single, backgroundcolor=\color{blue!5}]
$ python run_care_training.py
\end{lstlisting}
\noindent To reproduce the results presented in~\S\ref{sec:vefification}, we provide several functions in the {\texTTT{care\_rootcause\_viz.py}} script.  

\subsubsection{E2: Transferability Experiment}
We reuse the causal model constructed from the \textit{Husky}
simulator to determine the root cause of the functional faults in the \textit{Turtlebot-3} physical robot. The following command runs the experiment:
\begin{lstlisting}[basicstyle=\ttfamily\scriptsize, frame=single, backgroundcolor=\color{blue!5}]
$ python run_care_inference.py
\end{lstlisting}
\noindent The list of root causes for different ranks are printed in the terminal along with the accuracy, precision, and recall. To reproduce the results  presented in~\S\ref{sec:transferability}, we provide the {\texTTT{care\_transferibility\_viz.py}} script that produces the \textit{RMSE} plot in the {\texTTT{./care/fig}} directory.

\subsection{Using \textsc{CaRE} with external data}
\textsc{CaRE} can be applied to a different robotic system given the observational data as a {\texTTT{pandas.Dataframe}}. For example, update the {\texTTT{run\_care\_training.py}} script as follows:
%%%%%%%%code%%%%%%%%%%
%New colors defined below
\definecolor{codegreen}{rgb}{0,0.6,0}
\definecolor{codegray}{rgb}{0.5,0.5,0.5}
\definecolor{codepurple}{rgb}{0.58,0,0.82}
\definecolor{backcolour}{rgb}{1.0, 0.98, 0.98}
%Code listing style named "mystyle"
\lstdefinestyle{mystyle}{
  %backgroundcolor=\color{backcolour},
  commentstyle=\color{codegreen},
  keywordstyle=\color{magenta},
  numberstyle=\tiny\color{codegray},
  stringstyle=\color{codepurple},
  basicstyle=\ttfamily\footnotesize,
  breakatwhitespace=false,         
  breaklines=true,                 
  captionpos=b,                    
  keepspaces=true,                 
  %numbers=left,                    
  numbersep=5pt,                  
  showspaces=false,                
  showstringspaces=false,
  showtabs=false,                  
  tabsize=2
}
%"mystyle" code listing set
\lstset{style=mystyle}
\hypersetup{
    colorlinks=true,
    linkcolor=blue,
    filecolor=magenta,      
    urlcolor=blue,
    pdftitle={Overleaf Example},
    pdfpagemode=FullScreen,
    }
\begin{lstlisting}[language=python, frame=single,backgroundcolor=\color{blue!5}]
# read the observational data
df = pd.read_csv('observational_data.csv')
# read all columns
columns = df.columns
# Manipulable variables (configuration options)
manipulable_variables = ['option_1','option_2']
# Non-manipulable variables (performance metrics)
non_manipulable_variables = ['metric_1','metric_2']
# Performance objective (energy)
perf_objective = ['objective_1','objective_2']
\end{lstlisting}

\noindent Additionally, we added \href{https://github.com/softsys4ai/Reval/blob/husky/src/benchmark/README.md#cahnging-configuration-options}{\color{blue!80}instructions} to specify values for configuration options, change the experimental environment, and define a mission specification during observational data collection.

\section*{APPENDIX}
\subsection{Configuration Options in ROS {\texTTT{nav core}}} \label{sec:config_options}
\begin{table}[!ht]
%\vspace{-1em}
\centering
\caption{Configuration options in {\texTTT{base local planner}}}
\resizebox{\linewidth}{!}{%
\begin{tabular}{cll} 
\hline
\textbf{Parameters} & \textbf{Configuration options} & \textbf{Values/Range} \\ 
\hline
\multirow{11}{*}{Robot Configuration} & \texTTT{acc\_lim\_x} & 0 - 5 \\
 & \texTTT{acc\_lim\_y} & 0 - 5 \\
 & \texTTT{acc\_lim\_theta} & 0 - 6 \\
 & \texTTT{max\_vel\_x} & 0 - 1 \\
 & \texTTT{min\_vel\_x} & -0.1 - 0.2 \\
 & \texTTT{max\_vel\_theta} & 0 - 1 \\
 & \texTTT{min\_vel\_theta} & -1 - 0 \\  
 & \texTTT{min\_in\_place\_theta} & 0 - 0.5 \\ 
 & \texTTT{escape\_vel} & -0.2 - 0 \\ 
 & \texTTT{holonomic\_robot} & true \\  
 & \texTTT{y\_vels} & -0.3 - 0.3 \\ 
\hline
\multirow{3}{*}{Goal Tolerance} & \texTTT{yaw\_goal\_tolerance} & 0.1 - 3 \\
 & \texTTT{xy\_goal\_tolerance} & 0.1 - 0.4 \\
 & \texTTT{latch\_xy\_goal\_tolerance} & true, false \\ 
\hline
\multirow{6}{*}{Forward Simulation} & \texTTT{sim\_time} & 1 - 2 \\
 & \texTTT{sim\_granularity} & 0.015 - 0.03 \\
 & \texTTT{vx\_samples} & 1 - 30 \\
 & \texTTT{vtheta\_samples} & 1 - 30 \\
 & \texTTT{controller\_frequency} & 10 - 20 \\ 
\hline
\multirow{8}{*}{Trajectory Scoring} & \texTTT{meter\_scoring } & true, false \\
 & \texTTT{pdist\_scale } & 0.1 - 1 \\
 & \texTTT{gdist\_scale} & 0.5 - 1.5 \\
 & \texTTT{occdist\_scale} & 0.01 - 0.05 \\
 & \texTTT{heading\_lookahead } & 0.325 \\
 & \texTTT{heading\_scoring } & true, false \\
 & \texTTT{heading\_scoring\_timestep } & 0.8  \\
 & \texTTT{dwa} & true, false \\
 & \texTTT{publish\_cost\_grid} & true, false \\ 
\hline
Oscillation Prevention & \texTTT{oscillation\_reset\_dist} & 0.05 \\ 
\hline
Global Plan & \texTTT{prune\_plan} & true, false \\
\hline
\end{tabular}
}
\vspace{-1em}
\end{table}

\begin{table}[!ht]
%\vspace{-1em}
\centering
\caption{Configuration options in {\texTTT{global planner}}}
\begin{tabular}{ll} 
\toprule
\textbf{Configuration options} & \textbf{Values} \\ 
\hline
\texTTT{allow\_unknown} & 0 \\
\texTTT{default\_tolerance} & false \\
\texTTT{use\_dijkstra} & true \\
\texTTT{use\_quadratic} & true \\
\texTTT{use\_grid\_path} & false \\
\texTTT{old\_navfn\_behavior} & false \\
\texTTT{lethal\_cost} & 253 \\
\texTTT{neutral\_cost} & 50 \\
\texTTT{cost\_factor} & 3 \\
\texTTT{publish\_potential} & true \\
\texTTT{orientation\_mode} & 0 \\
\texTTT{orientation\_window\_size} & 1 \\
\texTTT{outline\_map} & true \\
\bottomrule
\end{tabular}
%\vspace{-1em}
\end{table}

\begin{table}[!t]
\centering
\caption{Configuration options in {\texTTT{costmap 2d}}}
\begin{tabular}{ll} 
\toprule
\textbf{Configuration options} & \textbf{Values/Range} \\ 
\hline
\texTTT{footprint\_padding} & 0.01 \\
\texTTT{update\_frequency} & 4 - 7 \\
\texTTT{publish\_frequency} & 1 - 4 \\
\texTTT{transform\_tolerance} & 0.2 - 2 \\
\texTTT{resolution} & 0.05 \\
\texTTT{obstacle\_range} & 5.5 \\
\texTTT{raytrace\_range} & 6 \\
\texTTT{inflation\_radius} & 1 - 10 \\
\texTTT{cost\_scaling\_factor} & 1 - 20 \\
\texTTT{combination\_method} & true, false \\
\texTTT{stop\_time\_buffer} & 0.1 - 0.3 \\
\bottomrule
\end{tabular}
\vspace{3em}
\end{table}

\newpage

\subsection{Configuration Setting for Root Cause Verification}\label{sec:verify_root_cause}
\begin{table}[!ht]
\centering
\caption{Configuration setting for Energy}
\scalebox{0.8}{
\begin{tabular}{ll|ll|ll} 
\toprule
\multicolumn{2}{c|}{Rank 1} & \multicolumn{2}{c|}{Rank 3} & \multicolumn{2}{c}{Rank 4} \\ 
\hline
\textbf{~}Options & Values & Options & Values & Options & Values \\ 
\hline
Cost\_scaling\_factor & 10 & Cost\_scaling\_factor & 10 & \textbf{Cost\_scaling\_factor} & 2 - 20 \\
update\_frequency & 4 & update\_frequency & 4 & \textbf{update\_frequency} & 1 - 7 \\
publish\_frequency & 3 & publish\_frequency & 3 & publish\_frequency & 3 \\
transform\_tolerance & 0.5 & \textbf{transform\_tolerance} & 0.2 - 2 & transform\_tolerance & 0.5 \\
combination\_method & 0 & \textbf{combination\_method} & 0, 1 & combination\_method & 0 \\
pdist\_scale & 0.75 & pdist\_scale & 0.75 & pdist\_scale & 0.75 \\
\textbf{gdist\_scale} & 0.5 - 4 & gdist\_scale & 1 & gdist\_scale & 1 \\
\textbf{occdist\_scale} & 0.01 - 2 & occdist\_scale & 0.1 & occdist\_scale & 0.1 \\
stop\_time\_buffer & 0.2 & stop\_time\_buffer & 0.2 & stop\_time\_buffer & 0.2 \\
yaw\_goal\_tolerance & 0.1 & yaw\_goal\_tolerance & 0.1 & yaw\_goal\_tolerance & 0.1 \\
xy\_goal\_tolerance & 0.2 & xy\_goal\_tolerance & 0.2 & xy\_goal\_tolerance & 0.2 \\
min\_vel\_x & 0 & min\_vel\_x & 0 & min\_vel\_x & 0 \\
\bottomrule
\end{tabular}
}
\end{table}

\begin{table}[!ht]
\centering
\caption{Configuration setting for Mission Success}
\scalebox{0.78}{
\begin{tabular}{ll|ll|ll} 
\toprule
\multicolumn{2}{c|}{Rank 1} & \multicolumn{2}{c|}{Rank 3} & \multicolumn{2}{c}{Rank 4} \\ 
\hline
\textbf{~}Options & Values & Options & Values & Options & Values \\ 
\hline
Cost\_scaling\_factor & 10 & Cost\_scaling\_factor & 10 & Cost\_scaling\_factor & 10 \\
update\_frequency & 4 & update\_frequency & 4 & update\_frequency & 4 \\
publish\_frequency & 3 & publish\_frequency & 3 & publish\_frequency & 3 \\
transform\_tolerance & 0.5 & \textbf{transform\_tolerance} & 0.2 - 2 & transform\_tolerance & 0.5 \\
combination\_method & 0 & combination\_method & 0 & \textbf{combination\_method} & 0, 1 \\
pdist\_scale & 0.75 & pdist\_scale & 0.75 & pdist\_scale & 0.75 \\
gdist\_scale & 1 & \textbf{gdist\_scale} & 0.5 - 4 & gdist\_scale & 1 \\
\textbf{occdist\_scale} & 0.01 - 2 & occdist\_scale & 0.1 & occdist\_scale & 0.1 \\
stop\_time\_buffer & 0.2 & stop\_time\_buffer & 0.2 & stop\_time\_buffer & 0.2 \\
yaw\_goal\_tolerance & 0.1 & yaw\_goal\_tolerance & 0.1 & \textbf{yaw\_goal\_tolerance} & 0.05 -1 \\
\textbf{xy\_goal\_tolerance} & 0.01 - 1 & xy\_goal\_tolerance & 0.2 & xy\_goal\_tolerance & 0.2 \\
min\_vel\_x & 0 & min\_vel\_x & 0 & min\_vel\_x & 0 \\
\bottomrule
\end{tabular}
}
\end{table}

\end{document}